\pdfoutput=1
\documentclass[11pt]{article}
\usepackage{acl}

\usepackage{times}
\usepackage{latexsym}

\usepackage[T1]{fontenc}

\usepackage[utf8]{inputenc}

\usepackage{microtype}

\usepackage{inconsolata}

\usepackage{xcolor}
\usepackage{graphicx}
\usepackage{booktabs}
\usepackage{url}
\usepackage{comment}

\usepackage{amsmath}
\usepackage{array}
\usepackage{amssymb}
\usepackage{cleveref}
\usepackage{makecell}
\usepackage{enumitem}
\usepackage{tipa}

\usepackage{xspace,mfirstuc,tabulary}
\usepackage{soul}
\usepackage{linguex}

\definecolor{darkyellow}{RGB}{235, 199, 19}

\newcommand{\CR}[1]{#1}

\usepackage{xpatch}
\usepackage{unravel}
\unravelsetup{max-action=1000, max-input=1000, max-output=1000}
\long\def\beginunravel#1\endunravel{\unravel{#1}}
\providecommand\beginunravel{}
\def\endunravel{}

\makeatletter

\xpretocmd\@declaredcolor
  {\my@hack@definetempcolor{#1}}
  {}{\PatchFailed}

\def\my@hack@color{\xglobal\definecolor}

\xpretocmd\XC@col@rlet
  {\my@hack@definetempcolor{#4}}
  {}{\PatchFailed}

\def\my@hack@definetempcolor#1{%
  \expanded{\my@hack@@definetempcolor{\zap@space#1 \@empty}}}%
\protected\def\my@hack@@definetempcolor#1{%
  \in@|{#1}%
  \ifin@
    \@ifundefinedcolor{#1}{\my@hack@splitcolor#1\@nil{\xglobal\definecolor}}{}%
  \fi
}

\newcommand\blfootnote[1]{%
  \begingroup
  \renewcommand\thefootnote{}\footnote{#1}%
  \addtocounter{footnote}{-1}%
  \endgroup
}

\def\my@hack@splitcolor#1|#2\@nil#3{#3{#1|#2}{#1}{#2}}
\makeatother

\definecolor{ambred}{HTML}{d7242b}
\definecolor{gpdandelion}{HTML}{f1ab32}
\definecolor{nongpcerulean}{HTML}{0097ce}

\newcommand{\amb}[1]{\textcolor{ambred}{#1}}
\newcommand{\gp}[1]{\textcolor{gpdandelion}{#1}}
\newcommand{\nongp}[1]{\textcolor{nongpcerulean}{#1}}

\title{Incremental Sentence Processing Mechanisms in Autoregressive Transformer Language Models}

\author{
Michael Hanna$^*$\\ILLC, University of Amsterdam\\\texttt{m.w.hanna@uva.nl}
\And
Aaron Mueller$^*$\\Northeastern University \& Technion -- IIT\\\texttt{aa.mueller@northeastern.edu}
}

\date{}

\begin{document}
\maketitle
\begin{abstract}
Autoregressive transformer language models (LMs) possess strong syntactic abilities, often successfully handling phenomena from agreement to NPI licensing. However, the features they use to incrementally process language inputs are not well understood. In this paper, we fill this gap by studying the mechanisms underlying garden path sentence processing in LMs. We ask: (1) Do LMs use syntactic features or shallow heuristics to perform incremental sentence processing? (2) Do LMs represent only one potential interpretation, or multiple? and (3) Do LMs reanalyze or repair their initial incorrect representations? To address these questions, we use sparse autoencoders to identify interpretable features that determine which continuation---and thus which reading---of a garden path sentence the LM prefers. We find that while many important features relate to syntactic structure, some reflect syntactically irrelevant heuristics. Moreover, while most active features correspond to one reading of the sentence, some features correspond to the other, suggesting that LMs assign weight to both possibilities simultaneously. Finally, LMs do not re-use features from garden path sentence processing to answer follow-up questions.\blfootnote{\hspace{-0.45em}$^*$Equal contribution.}\footnote{Code and data are available at \url{https://github.com/hannamw/GP-mechanisms/}.}
\end{abstract}

\section{Introduction}
Syntactic ambiguities abound in natural language. %
For example, given the fragment ``After the woman drank the water\ldots'', \textit{the water} could be either the object of \emph{drank} (in which case one could end the sentence here), or the subject of the main clause (in which case ``was all gone'' would be a valid continuation). Despite LMs' impressive performance on syntactic tasks \citep{hu-etal-2020-systematic}, the mechanisms that underlie their processing of syntactic structure---and temporary ambiguities therein---are not well understood. 
Past work has found LM attention heads dedicated to processing certain syntactic relations \citep{vig-belinkov-2019-analyzing} and used LMs' representational structure to predict dependency relations \citep{hewitt-manning-2019-structural};
nonetheless, these results only show that structural information can be extracted from LM representations---and not that these representations are causally implicated in LM processing. It thus remains unclear whether LMs rely on structure-related features, represent the multiple possible completions to an incomplete ambiguous utterance, or revise representations in light of new disambiguating evidence.

In the psycholinguistics literature, similar questions have been studied in humans using \textbf{garden path sentences}, which initially appear to have one structure, but which are later revealed to have another. When humans encounter the unexpected resolution of these sentences, their reading is delayed. Different theories of human sentence processing predict different delays; by recording reading times on carefully designed test materials, one can thus empirically test such theories \citep{lewis2000falsifying,gibson2000distinguishing}. While prior work on LMs has used garden path sentences as a testbed for the psychometric fit of LM surprisals to predict human reading times \citep{van2021single,arehalli-etal-2022-syntactic,huang2024large}, we propose to instead use them to understand how LMs incrementally process sentences.

\begin{figure*}[t]
    \centering
    \includegraphics[width=0.95\linewidth]{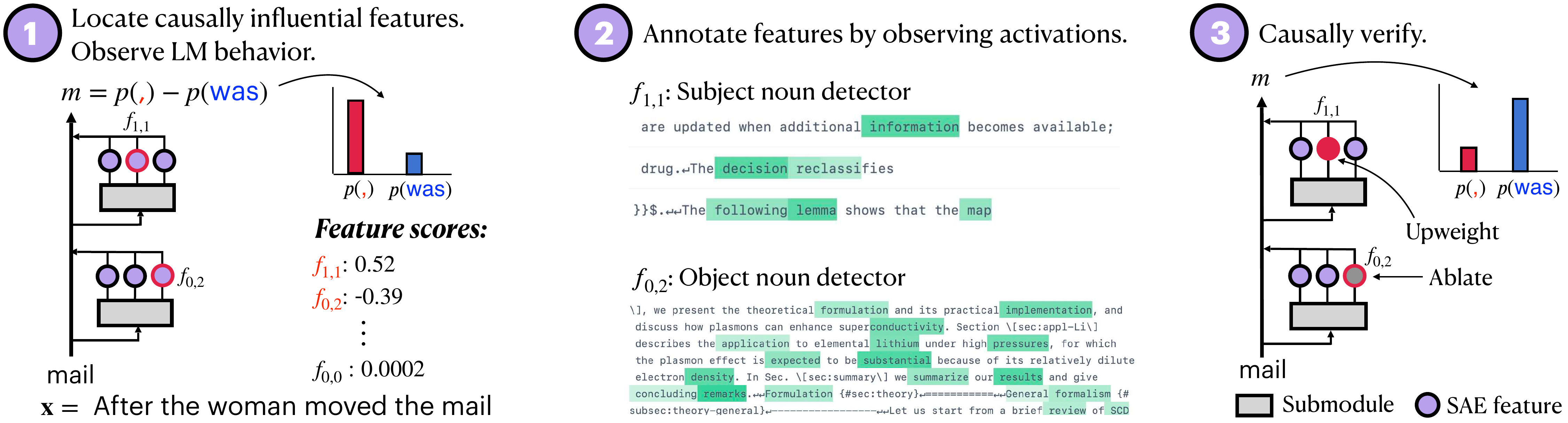}
    \caption{\textbf{Overview.} We use sparse autoencoders to decompose model activations into a discrete set of human-interpretable components (features). We score each feature by its causal contribution to continuations associated with each reading of a garden path sentence. We manually interpret the top-scoring features and causally verify their functional role in the network by targetedly up- or downweighting them to change the model's preferred reading.}
    \label{fig:overview}
\end{figure*}

In this study, we present a mechanistic investigation of how LMs incrementally process sentences and how they handle temporary ambiguities using garden path (GP) sentences as a case study. Using sparse autoencoders and causal interpretability methods, we uncover the causally relevant features (and mechanisms composed thereof) that explain why LMs assign higher probabilities to particular completions. With these methods, we investigate 3 research questions (RQs), and find the following:

\textbf{RQ1}: \emph{Do LMs use syntactic features or spurious heuristics to incrementally process sentences?} Many of the most important features LMs use are interpretable and syntax-related; however, some uninterpretable or spurious features exist.

\textbf{RQ2}: \emph{Do LMs hold on to multiple interpretations of the sentence simultaneously, or commit to the most likely one?} LMs' representations encode multiple interpretations.

\textbf{RQ3}: \emph{Given disambiguating evidence, do LMs repair or reanalyze their initial structural predictions?} LMs do not repair or rely on their prior structural predictions; however, they also do not generate new structural features via reanalysis.

\section{Background}
\subsection{Incremental Sentence Processing}\label{subsec:incremental}
Many linguistic theories posit that humans parse their linguistic input, mapping from sentences to a representation with information about its structure 
\citep{vanGompel2007syntactic}. We do so \emph{incrementally}, building up representations prior to the end of the sentence \citep{marslenwilson1975sentence}.

How humans perform incremental parsing is hotly debated. Of particular interest is how we handle the fact that partial sentences often have multiple valid parses \citep{fodor1974psychology}. Do we parse sentences serially, considering one parse at a time \citep{frazier1979comprehending,fodor1998reanalysis}, or in parallel, considering many at once \citep{gorrell1987studies,gibson1991computational,jurafsky1996probabilistic}? And upon encountering evidence that rules out specific parses, do we repair our representations \citep{lewis1998reanalysis}, or reanalyze the input entirely \citep{grodner2003against}?

Psycholinguists often test theories of incremental parsing with \emph{garden path sentences}, which suggest one parse, but ultimately have another. Consider the incomplete sentence ``The guitarist knew the song...''. A reader could either interpret \emph{song} as an object of the verb \emph{knew}, or the subject of a sentential clause (i.e., ``The guitarist knew (that) the song...''). A period would be a valid completion in the former case but not the latter, where a verb phrase like ``...was too long'' would be more fitting. Most readers find the first reading more likely, so observing a completion consistent with the second typically results in significant spikes in reading times \citep{frazier1987sentence}.

\subsection{Sentence Processing in LMs}\label{sec:sentence-processing-lms}
How LMs process and represent sentences is similarly well-studied. Work on structural probes has attempted to reconstruct parses from LM representations using learned similarity functions or probes \citep{hewitt-manning-2019-structural,hall-maudslay-etal-2020-tale,white-etal-2021-non,arps-etal-2022-probing}. Others have found attention heads whose attention corresponds to syntactic relations, though no general parsing head exists \citep{vig-belinkov-2019-analyzing,clark-etal-2019-bert,htut2019attention}. Researchers have also trained probes to extract features like coreference relations or part of speech from LM representations \citep{tenney2018what,jawahar-etal-2019-bert}.

However, these analyses can yield only limited insights into LMs' incremental processing mechanisms. 
Most study LMs with bidirectional attention, which do not perform \emph{incremental} sentence processing. Moreover, few causally verify their mechanisms' relevance to model processing, even though  probes often capture functionally irrelevant information \citep{ravichander-etal-2021-probing,elazar-etal-2021-amnesic}. While causal techniques have been used in other settings 
\citep{vig2020investigating,finlayson-etal-2021-causal,lasri-etal-2022-probing}, they have rarely been applied to questions of ambiguity in syntactic structure and incremental processing; \citet{eisape-etal-2022-probing} do so, but using a probe which assumes a specific mechanism unlikely to be encoded by the LM itself.

With this in mind, we use garden path sentences as a case study in LMs' incremental sentence processing mechanisms. Prior work using LM behavior on such sentences to model human reading times \citep{van2018modeling,wilcox-etal-2021-targeted,arehalli-etal-2022-syntactic,oh-schuler-2023-surprisal} finds that LMs do exhibit garden-path effects, though they underpredict human effects. Less work has used garden path sentences to observe \emph{how} LMs arrive at these probabilities and surprisals. \citet{li2024incremental} attempt this, but study masked LMs without causal methods. We ask: how can we find and causally verify the mechanisms that LMs use to incrementally process garden path sentences?

\subsection{Locating Interpretable Mechanisms with Sparse Feature Circuits}\label{sec:SAEs}

\CR{To understand how LMs incrementally process sentences, we must understand the features that they use to represent their input. Earlier work did so by studying individual LM neurons \citep{sajjad-etal-2022-neuron} and searching for the inputs that cause them to activate the most strongly. A neuron that activates on the subjects of sentences might thus be inferred to implement a subjecthood feature.}

\CR{However, feature representations in neural networks are often distributed \citep{hinton1986distributed,smolensky1986distributed}. Neurons are thus often \emph{polysemantic}, representing multiple unrelated features at once, which makes them challenging to interpret \citep{olah2017feature,elhage2022toymodelssuperposition}; for example, \citet{bolukbasi2021interpretability} find a neuron that activates on sentences about song meanings, objects in containers, and historical dates.}

We thus opt to interpret the features of sparse autoencoders (SAEs; \citealp{bricken2023monosemanticity}), autoencoders trained on the output activations of LM submodules. Let $\mathbf{x}$ be the submodule's output activation; the SAE computes
\begin{align}
    \mathbf{f} &= \text{ReLU}(W_e(\mathbf{x} - \mathbf{b}_d) + \mathbf{b}_e)\\
    \mathbf{\hat{x}} &= W_d\mathbf{f} + \mathbf{b}_d,
\end{align}
where $\mathbf{f}$ is the feature vector, and $\mathbf{\hat{x}}$ is the reconstructed activations. Henceforth, we refer to a single dimension of $\mathbf{f}$ as a \emph{feature}. SAEs are trained to reconstruct $\mathbf{x}$ with sparse regularization on $\mathbf{f}$; the regularizer and bias terms lead a feature's activation to be non-$0$ only when it causes parts of $\mathbf{x}$ to differ from their mean value. This makes SAE features more \emph{monosemantic} than LM neurons, and therefore more interpretable. 

\CR{In order to assemble SAE features into entire model mechanisms, we use \emph{circuit} analysis \citep{olah2020zoom}. A circuit is the minimal subset of the language model’s computation graph that recovers the whole LM's performance on a given task \citep{wang2023interpretability,conmy2023towards,hanna2023gpt2}. In this case, each node in the graph is an SAE feature, and each edge represents a cause-effect relationship. The first node in the graph is the embeddings, the final node is the model’s output logits, and all intermediate nodes are features from SAEs trained on neurons from the model's residual stream or attention head / MLP outputs.} 

\CR{Circuits can be conceptualized as a causal abstraction of the language model. If a node is in the graph, it is causally relevant to the LM's task abilities. Moreover, if a node has an edge to another, this implies that the activation of the second crucially depends on the activation of the first.}

We follow \citeposs{marks2024sparse} approach to finding such \emph{sparse feature circuits}. We say a feature $f$ is causally relevant if, given a metric $m$ that measures the language model's behavior, setting $f$'s value to 0 causes a large change in $m$;\footnote{While setting neurons to 0 is unprincipled, zero-ablating sparse features is not, as feature activations are non-$0$ only when they cause parts of $\mathbf{x}$ to differ from their mean value.} the magnitude of this change is $f$'s \textbf{indirect effect} (IE; \citealp{pearl2001indirect}) on $m$. We aim to find features with high IE.

Computing each feature's IE is expensive, so we compute a linear approximation $\hat{\text{IE}}$. \CR{Attribution patching (AtP; \citealp{nanda2023attribution}) is one such approximation, estimating the indirect effect $\hat{\text{IE}}$ of a feature with activation $a$ on input $x$ as
\begin{equation}
\hat{\text{IE}} = a\cdot \frac{\partial m}{\partial a}\Big|_{x}.
\label{eq:atp}
\end{equation}
$\frac{\partial m}{\partial a}$ is computed by backpropagation from $m$. Conceptually, the slope of the metric $m$ with respect to the feature's activation $a$ is multiplied by the change in the feature's activation upon being zeroed ($a - 0$, which simplifies to $a$). In practice, AtP can often be inaccurate, so we use the improvement of \citet{marks2024sparse}: attribution patching with integrated gradients (AtP-IG). Inspired by integrated gradients for input attribution \citep{sundarajan201axiomatic}, AtP-IG computes an average $\frac{\partial m}{\partial a}$ across several intermediate activations of $f$ between $a$ and 0, leading to better estimates of the IE. See App.~\ref{app:atp-ig} for details on AtP-IG.}
 
After estimating the $\hat{\text{IE}}$ of each feature, we select all features and edges whose $\hat{\text{IE}}$ is over a chosen threshold; this yields a circuit. By verifying that $m$'s value remains the same when the features outside our circuit are ablated, we verify that the mechanism captured by the circuit is faithful to that of the full model.

\section{Models}
To analyze incremental parsing in LMs, we must study autoregressive LMs.\footnote{Masked LMs often have strong syntactic abilities \citep{goldberg2019assessingbertssyntacticabilities} but receive the left \emph{and} right context of each token, invalidating them as models of incremental processing.} We analyze Pythia-70m-deduped \citep{biderman2023pythia}, and Gemma-2-2b \citep{gemmateam2024gemma2improvingopen}, as these have publicly available SAEs. We focus primarily on Pythia-70m in the main text due to its smaller size; results for Gemma-2-2b are in App.~\ref{app:other-models}.

\section{Do LMs use syntactic features to process garden path sentences?}

\CR{In this study, we use garden path sentences to investigate incremental processing in LMs. Garden path sentences are useful objects of study because they contain temporary structural ambiguities that are eventually resolved. The sentences we study must be ambiguous such that we may determine if one or many possible readings are represented (RQ2). Moreover, this ambiguity must eventually be resolved such that we may study how incorrect representations might be handled (RQ3). Thus, garden path sentences are ideal stimuli with which to answer our RQs; indeed, they are often used for this purpose in psycholinguistics (\S\ref{subsec:incremental}).}

\subsection{Behavioral Analysis}\label{subsec:behavioral}
Before finding the features that underlie LM garden path sentence processing, we first verify that the LMs we study exhibit garden path effects.  

\paragraph{Dataset} We probe LMs' readings of garden path sentences using an adaptation of \citeposs{arehalli-etal-2022-syntactic} dataset of 72 garden path sentences. This contains 3 structures (NP/Z, NP/S, and MV/RR) with 24 sentences each. Each structure name refers to the garden path/actual interpretations of the sentence's ambiguous material, respectively. For example, in \Cref{tab:examples} NP/Z, ``signed'' could take either an NP complement (``the bill'') or a zero complement. In \Cref{tab:examples} NP/S, ``the song'' could be the NP complement of ``knew'' or the start of a sentential complement. Finally, in \Cref{tab:examples} MV/RR, ``brought'' could be the main verb or part of a reduced relative clause.

\begin{table}[t]
\begin{centering}
    \resizebox{\linewidth}{!}{
    \begin{tabular}{l>{\raggedright}p{5.9cm}cc}
    \toprule
    Structure & Example Sentence & GP & Non-GP\\
    \midrule
    NP/Z &After the politician \amb{signed}/\gp{rejected}/\nongp{arrived} the bill&,&was\\%, had, would, should, might\\
    NP/S &The guitarist \amb{knew}/\gp{wrote}/\nongp{said} the song&.&was\\%, had, would, should, might\\
    MV/RR &The woman \amb{brought}/\gp{moved}/\nongp{shown}\newline{}the mail&.&was\\%, had, would, should, might\\
    \bottomrule
    \end{tabular}}
    \caption{Examples from our dataset, adapted from \citet{arehalli-etal-2022-syntactic}. For each sentence, inserting the \gp{yellow} word makes it compatible with only \gp{garden path (GP)} continuations; the \nongp{blue} word permits only \nongp{non-garden-path (Non-GP)} continuations. The \amb{red} words leave it \amb{ambiguous}, compatible with either.}\label{tab:examples}
\end{centering}
\end{table}

For each ambiguous sentence, we craft two unambiguous versions, which permit only one reading. For example, in \Cref{tab:examples} NP/Z, we replace the ambitransitive ``signed'', with the strictly transitive ``rejected'' (forcing the garden-path reading) or the intransitive ``arrived'' (forcing the opposite).

\begin{figure}
    \centering
    \includegraphics[width=\linewidth]{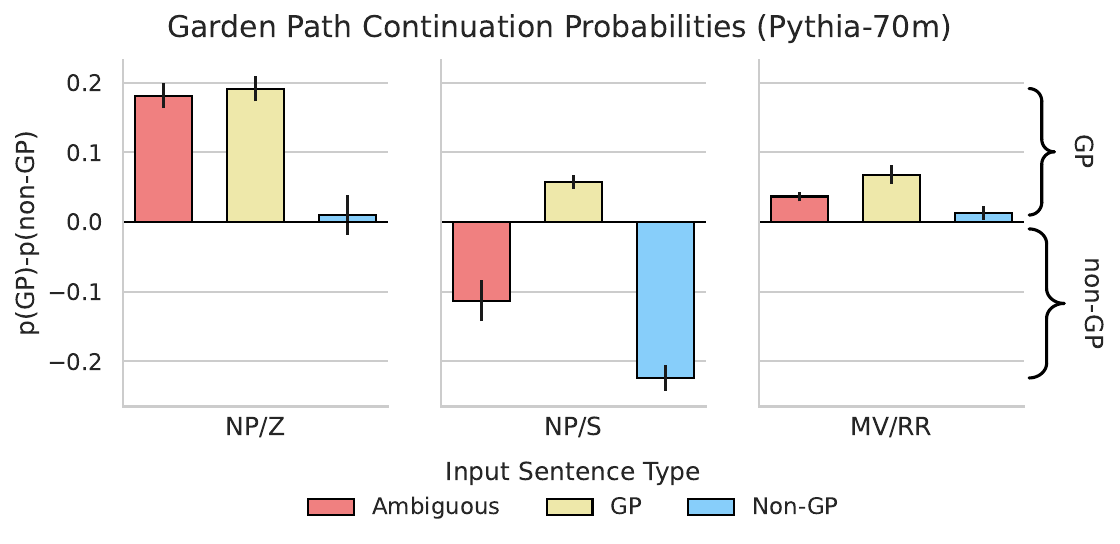}
    \caption{Mean difference in probability of tokens corresponding to garden path (``,''/``.'') and non-garden-path (``was'') readings of the input for Pythia-70m, grouped by garden path structure. Error bars indicate the standard error of the mean. Inputs are either \amb{ambiguous}, or compatible with only a \gp{garden-path} or \nongp{non-garden-path} reading. GP tokens are more likely given GP inputs; non-GP are more likely with non-GP inputs. In ambiguous cases, Pythia-70m prefers the GP reading, except for on NP/S.}
    \label{fig:behavioral_pythia-70m}
\end{figure}

\paragraph{Experiment} For each sentence, we record the probability given by the LM to next tokens consistent with the garden path or non-garden-path reading; we denote these $p($GP$)$ and $p($non-GP$)$, respectively. For NP/Z sentences, we define $p($GP$)$ as $p($,$)$; for NP/S and MV/RR, we define $p($GP$)$ as $p($.$)$. For all sentence structures, we define $p($non-GP$)$ as
the probability of `` was''. This roughly measures the LM's reading of the sentence: continuing ``After the politician signed the bill'' with a comma implies that ``signed'' took ``the bill'' as a complement, as in the GP reading. Continuing it with `` was'' implies that ``signed'' took no complement, as in the non-GP reading.

\paragraph{Results} Our results (\Cref{fig:behavioral_pythia-70m}) show that Pythia-70m correctly up- and downweights garden path tokens in contexts that do and do not license them. Given GP inputs, the model gives more probability to GP tokens, and less to non-GP tokens, compared to when it receives ambiguous inputs; this trend holds across garden path structures. For non-GP inputs, this trend is reversed.

For ambiguous inputs, the model prefers garden-path continuations in the NP/Z and MV/RR cases, but non-garden-path in the NP/S case. This agrees with prior evidence from both humans and LMs showing lower reading times and surprisals for non-garden-path continuations to inputs with NP/S ambiguities, compared to NP/Z \citep{grodner2003against,sturt1999structural,van2018modeling}.

\paragraph{Discussion} In the NP/Z and MV/RR cases, non-GP inputs only manage to reduce the model's bias for GP continuations to near 0, not eliminate it. We hypothesize that this has two causes. First, although NP/Z sentences are common objects of study in the psycholinguistics literature \CR{\citep{christianson-2001-linger,christianson2006younger}}, their non-GP readings are somewhat unnatural; this also applies to our non-GP versions. In normal text, these sentences would include a comma after the verb if the non-GP reading were intended (and models \emph{do} prefer the non-GP reading given a comma). 

\CR{Second, our operationalization of $p($GP$)$ and $p($non-GP$)$ has limitations. For non-GP MV/RR sentences, $p(\text{,})$ and $p(\text{was})$ are both low; the model gives the most probability to \emph{to}, which does not definitively distinguish between the two readings. For non-GP NP/Z sentences, $p(\text{,})$ and $p(\text{was})$ are higher, but measuring $p(\text{was})$ alone may miss much of the probability assigned to non-GP continuations more generally. Ideally, we would measure the probability of all full GP and non-GP-implying continuations (which might span multiple tokens), rather than measuring the probability two single next tokens.} Unfortunately, this is computationally infeasible, but see App.~\ref{app:behavioral} for more discussion of this issue and \S\ref{subsec:probes} for another way to determine an LM's reading of ambiguous input, yielding similar results. Because MV/RR sentences have low $p($GP$)$ and $p($non-GP$)$, we exclude them from all following analyses.\footnote{See \Cref{app:probe-unambiguous} for a non-behavioral way to measure models' readings of these sentences.}

\subsection{Feature Circuit Analysis}\label{subsec:feature}
Now, we identify and analyze the feature circuits responsible for Pythia-70m's garden path effects.

\paragraph{Experiment} We investigate circuits composed of causally relevant features from the Pythia-70m SAEs of \citet{marks2024sparse}; these SAEs have 32,768 features each. We use AtP-IG (\S\ref{sec:SAEs}) to find the features that most influence the difference in probabilities assigned to garden-path and non-garden-path continuations of ambiguous sentences, $m=p(\text{GP})$ - $p(\text{non-GP})$. We keep features with $\hat{\text{IE}}$ > 0.1, and edges with $\hat{\text{IE}}$ > 0.001. We then manually annotate each feature in the circuit, using Neuronpedia \citep{neuronpedia} to visualize feature activations on text from The Pile \citep{gao2020pile} on which each feature activates strongest.\footnote{Features can be viewed online via Neuronpedia at \url{https://www.neuronpedia.org/pythia-70m-deduped/}}

\begin{figure}
\centering
\includegraphics[width=0.9\linewidth]{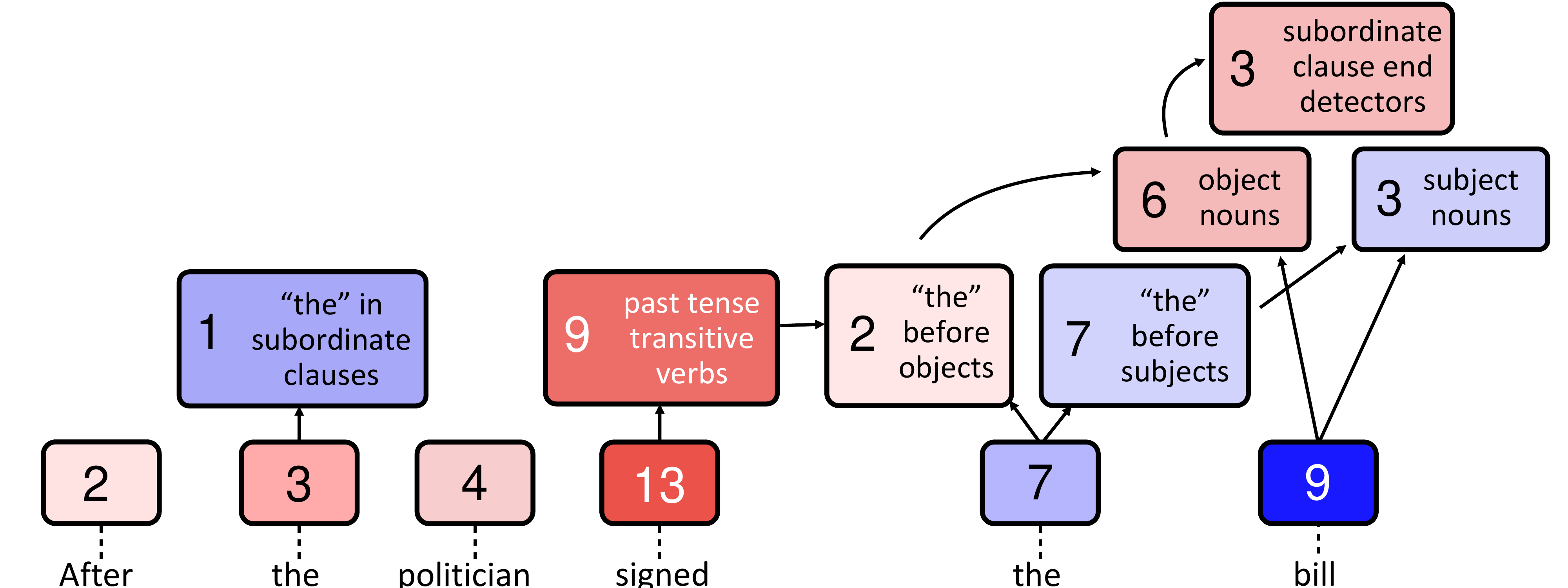}
\caption{Pythia-70m's feature circuit for processing NP/Z garden path sentences. We group features by their functional role in the circuit and display the number of features in each group. Red features have negative scores and promote the garden path reading; blue features, with positive scores, do the opposite. Unlabeled early-layer features are word detectors. Many late-layer features encode syntactic features, whereas early-layer features largely consist of word detectors and heuristics.} \label{fig:circuit-diagram}
\end{figure}

\begin{table*}
    \centering
    \resizebox{0.9\linewidth}{!}{
    \begin{tabular}{llc}
    \toprule
    Feature & Activates on & Examples\\
    \midrule
    0/8234 & \makecell{the word \emph{the}} & \makecell{\sethlcolor{white}\textcolor{black}{\hl{Since 2001,}}\sethlcolor{RGB|22,22,255}\textcolor{white}{\hl{ the}}\sethlcolor{white}\textcolor{black}{\hl{ variant commonly in use is}}\sethlcolor{RGB|69,69,255}\textcolor{white}{\hl{ the}}\sethlcolor{white}\textcolor{black}{\hl{ Category 5e specification}}\\
    \sethlcolor{white}\textcolor{black}{\hl{On September 26, 2006}}\sethlcolor{RGB|0,0,255}\textcolor{white}{\hl{ the}}\sethlcolor{white}\textcolor{black}{\hl{ University of Phoenix acquired}}\sethlcolor{RGB|102,102,255}\textcolor{white}{\hl{ the}}\sethlcolor{white}\textcolor{black}{\hl{ naming}}}\\
    \midrule %
    4/14907 & \makecell{ends of\\sub. clauses} &\makecell{\sethlcolor{white}\textcolor{black}{\hl{Finally, after}}\sethlcolor{RGB|223,223,255}\textcolor{black}{\hl{ years}}\sethlcolor{white}\textcolor{black}{\hl{ of}}\sethlcolor{RGB|222,222,255}\textcolor{black}{\hl{ watching}}\sethlcolor{RGB|175,175,255}\textcolor{black}{\hl{ youtube}}\sethlcolor{RGB|134,134,255}\textcolor{black}{\hl{ videos}}\sethlcolor{white}\textcolor{black}{\hl{ on}}\sethlcolor{RGB|236,236,255}\textcolor{black}{\hl{ that}}\sethlcolor{RGB|66,66,255}\textcolor{white}{\hl{ topic}}\sethlcolor{RGB|208,208,255}\textcolor{black}{\hl{,}}\sethlcolor{white}\textcolor{black}{\hl{ I made}}\\
    \sethlcolor{white}\textcolor{black}{\hl{When it}}\sethlcolor{RGB|214,214,255}\textcolor{black}{\hl{ released}}\sethlcolor{white}\textcolor{black}{\hl{ alongside}}\sethlcolor{RGB|239,239,255}\textcolor{black}{\hl{ Fire}}\sethlcolor{white}\textcolor{black}{\hl{ Em}}\sethlcolor{RGB|158,158,255}\textcolor{black}{\hl{blem}}\sethlcolor{white}\textcolor{black}{\hl{ F}}\sethlcolor{RGB|113,113,255}\textcolor{white}{\hl{ates}}\sethlcolor{RGB|223,223,255}\textcolor{black}{\hl{ in}}\sethlcolor{RGB|131,131,255}\textcolor{black}{\hl{ June}}\sethlcolor{RGB|217,217,255}\textcolor{black}{\hl{ of}}\sethlcolor{RGB|85,85,255}\textcolor{white}{\hl{ 2015}}\sethlcolor{RGB|240,240,255}\textcolor{black}{\hl{,}}\sethlcolor{white}\textcolor{black}{\hl{ Fire}}}\\
    \midrule %
    3/835 & \makecell{subjects of \\sent. clauses} &\makecell{\sethlcolor{white}\textcolor{black}{\hl{A hearing officer would determine if}}\sethlcolor{RGB|214,214,255}\textcolor{black}{\hl{ a}}\sethlcolor{RGB|37,37,255}\textcolor{white}{\hl{ complaint}}\sethlcolor{white}\textcolor{black}{\hl{ has merit, requiring}}\\
    \sethlcolor{white}\textcolor{black}{\hl{\ldots to learn how}}\sethlcolor{RGB|201,201,255}\textcolor{black}{\hl{ the}}\sethlcolor{RGB|130,130,255}\textcolor{black}{\hl{ United}}\sethlcolor{RGB|39,39,255}\textcolor{white}{\hl{ States}}\sethlcolor{RGB|178,178,255}\textcolor{black}{\hl{ and}}\sethlcolor{white}\textcolor{black}{\hl{ key}}\sethlcolor{RGB|98,98,255}\textcolor{white}{\hl{ players}}\sethlcolor{RGB|234,234,255}\textcolor{black}{\hl{ around}}\sethlcolor{white}\textcolor{black}{\hl{ the}}\sethlcolor{RGB|128,128,255}\textcolor{black}{\hl{ world}}}\\
    \midrule %
    4/8505  & \makecell{object nouns,\\nouns in PPs} &\makecell{\sethlcolor{white}\textcolor{black}{\hl{Justin Trudeau used the}}\sethlcolor{RGB|248,248,255}\textcolor{black}{\hl{ Canada}}\sethlcolor{RGB|184,184,255}\textcolor{black}{\hl{ Day}}\sethlcolor{RGB|0,0,255}\textcolor{white}{\hl{ celebrations}}\sethlcolor{RGB|201,201,255}\textcolor{black}{\hl{ in}}\sethlcolor{RGB|123,123,255}\textcolor{white}{\hl{ Ottawa}}\sethlcolor{white}\textcolor{black}{\hl{ to name}}\\
    \sethlcolor{white}\textcolor{black}{\hl{\ldots than for Alan Shepard. He left the}}\sethlcolor{RGB|24,24,255}\textcolor{white}{\hl{ hotel}}\sethlcolor{white}\textcolor{black}{\hl{ shortly}}\sethlcolor{RGB|241,241,255}\textcolor{black}{\hl{ after}}\sethlcolor{RGB|104,104,255}\textcolor{white}{\hl{ midnight}}}\\
    \bottomrule
    \end{tabular}}
    \caption{Interpretable residual stream features implicated in Pythia-70m's garden path sentence processing. We list each feature's layer and feature-index, as well as a description of what the feature activates highly on. Each example shows how strongly the feature activated on each token; darker highlighting indicates larger activations.}
    \label{tab:features}
\end{table*}

\paragraph{Results} Running AtP-IG yielded circuits containing 155 (NP/S) and 65 (NP/Z) sparse features; we manually annotate all of these.\footnote{Past work has used LMs as annotators \citep{bills2023language}, but we find them to be poor annotators of syntactic features.} To measure how well these features capture the behavior of the full model, we measure faithfulness, following the definition in \citet{marks2024sparse}. These circuits have faithfulness 0.20 (NP/S) and 3.48 (NP/Z).\footnote{Approaching a faithfulness of 1 requires including many hundreds of features for Pythia, and over 1000 for Gemma.} Significant deviations from 1.0 imply that there are important features we have not captured; thus, while we cannot claim to have annotated the full mechanism, we can nonetheless still analyze the most highly influential features, which provide sufficient evidence to address our RQs. See App.~\ref{app:faithfulness} for the metric definition, implementation details, and a deeper discussion of these faithfulness values.

\Cref{fig:circuit-diagram} displays a simplified circuit for NP/Z, where we manually group similar features together. The simplified NP/S circuit and the oversized full circuits are in App.~\ref{app:full-feature-circuit}. 
We present selected features' activations on highly-activating sentences in Table~\ref{tab:features} to support our annotations.

\textbf{Features in lower layers often correspond to interpretable low-level features.} Many lower-layer features detect word-level attributes, rather than high-level sentence-specific syntactic information. The vast majority of our circuit's features are word detectors that activate only on one specific word, located in the model's embeddings or first two layers. For example, Feature 0/8234 activates only on the word \emph{the} (\Cref{tab:features}). Other features are slightly higher-level, activating on nouns or past tense verbs. Notably, while most such features have no obvious syntactic relation with either reading (e.g. the presence of the word ``the'' should be neutral with respect to which reading it suggests), they have a non-zero impact on the preferred reading.

\textbf{Higher layer features encode syntactic attributes relevant to garden path sentence processing.}
The features in Pythia-70m's upper layers often encode sentence-level syntactic information that distinguishes between different readings of garden path sentences. For example, the final layers of the model's circuit for NP/Z sentences (\Cref{fig:circuit-diagram}) include features that detect subjects, objects, and ends of subordinate clauses. Reading the final noun as an object and part of the subordinate clause corresponds to the GP reading; reading the final noun as a subject outside of the subordinate clause corresponds to the opposite. The scores assigned to features match their semantics: non-GP feature scores are positive; pro-GP are negative.

\Cref{tab:features} shows each feature's activations. Feature 4/14907, for example, detects ends of subordinate clauses; every position at which it activates \textit{could} be a valid end to the subordinate clause containing it, given no information about the following tokens. It precisely distinguishes the two readings of NP/Z sentences: in the garden path reading of ``After the politician signed the bill'', the clause might end at \textit{bill}, while in the non-GP reading, it ends at \textit{signed}. 

Feature 3/835 distinguishes the readings of NP/S sentences, activating on subjects of sentential complements. In an NP/S sentence such as ``The guitarist knew the song'', \textit{the song} can either be the object of \textit{knew} (the GP reading) or the subject of a new phrase (non-GP); this feature clearly corresponds to the latter reading. Finally, Feature 4/8505 activates primarily on object nouns and nouns in prepositional phrases. This corresponds not only to the GP reading in NP/Z and NP/S sentences, but also to the accusative case, hinting that the model may have learned a general linguistic concept.

\textbf{Some features are uninterpretable.} Although many SAE features are interpretable, some activate seemingly at random, or across almost all text. The latter could be interpreted as a prior, which always influences the model's prediction, but most have no clear interpretation. These features have a non-zero effect on model predictions, though their effect direction is inconsistent. We omit these features from our analysis, but we hope they will be interpretable as SAEs or interpretability methods improve. 

\subsection{Causal Analysis}\label{subsec:causal}
Though many high-importance features encode syntactic attributes, this is no guarantee that the model relies on them. To confirm this, we causally intervene on the discovered interpretable features, and verify that model output changes as we expect. See App.~\ref{app:large-dataset} for a large-scale version of this experiment.

\paragraph{Experiment} We focus on three groups of model features, which detect: 1) subjects, 2) objects, and 3) either ends of clauses (NP/Z) or sentential-clause verbs (NP/S). For NP/Z sentences, we attempt to induce the dispreferred non-GP reading by setting the subject detectors' activation to a high value (2.0) at the final noun of each sentence, while clamping object detectors off (to 0). End-of-clause detectors are set to 2.0 at the verb position, and 0 on the final noun. For NP/S sentences, we induce the GP reading: at the final noun, we set the subject and object detectors to 0 and 2.0 respectively; we also turn the sentential-clause-verb detectors off. As a control, we choose 3 groups of random features (equal in number to the original groups) to clamp on or off. In all settings, we intervene during the forward pass, and compute $p($GP$)$ and $p($non-GP$)$.

\begin{figure}
    \centering
    \includegraphics[width=\linewidth]{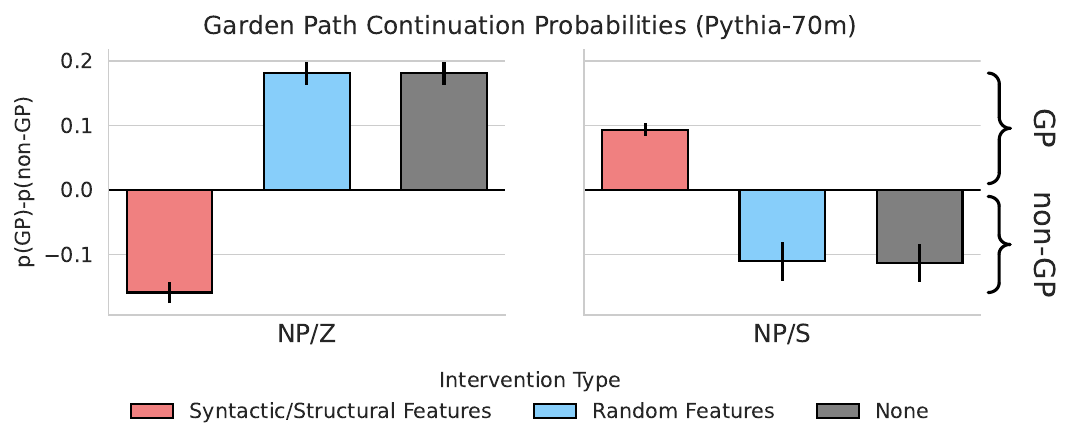}
    \caption{Mean difference in probability of GP and non-GP continuations under interventions for Pythia-70m. Error bars indicate the standard error of the mean.
    Interventions on interpretable features reverse model behavior, as expected; random interventions do nothing.}
    \label{fig:causal_pythia-70m}
\end{figure}

\paragraph{Results} Our results (\Cref{fig:causal_pythia-70m}) suggest that the features we find are causally relevant. Turning the subject and object features on and off respectively, and altering the end-of-clause features, reverses the model's typical preference for the GP reading in the NP/Z scenario; ablating the same number of randomly chosen features does nothing. The analogous NP/S intervention causes the model to prefer the GP reading, while performing no interventions or random ones yields the opposite.

\section{Do LMs consider one or multiple readings of garden path sentences?}
Here, we investigate whether LMs consider multiple readings of garden path sentences simultaneously. We reason that, although $p($GP$)$ and $p($non-GP$)$ are non-zero in all cases, the model may not explicitly represent both alternatives. \CR{We thus say that a model considers just one reading if, given an ambiguous input, it activates only features corresponding to one reading of the input, as opposed to multiple. A model considers multiple if it activates features that correspond to multiple readings of the input---e.g., if both subject \emph{and} object detectors fire on the final noun of an NP/Z or NP/S sentence.}

\subsection{Evidence from model feature analysis}
\paragraph{Experiment} We can test if LMs consider one or multiple readings of garden path sentences by checking if ambiguous inputs cause features corresponding to both readings to activate. We thus run the model on our ambiguous data and record the activations of the interpretable pro-GP and anti-GP features that we identified in layers 3-5 of the model, in \S\ref{subsec:feature}.
If the model only considers one reading, only features corresponding to one reading should activate; if features corresponding to both activate, we conclude that the model considers multiple. Recall that as features are inactive on almost all inputs, non-zero activations are meaningful.

\paragraph{Results} We find %
that in both the NP/Z and NP/S cases, pro- and non-GP features have non-zero average activations, ranging from 0.27 to 0.41. Similarly, the percent of features active is above 50\% for both categories, and both NP/Z and NP/S sentences. This suggests that models explicitly represent both readings of a garden path sentence.

\subsection{Evidence from structural probes}\label{subsec:probes}
We can also directly assess if the model considers both readings using structural probes (\S\ref{sec:sentence-processing-lms}), which map from LM representations to a distribution over parses of the LM's input. The two readings of NP/Z and NP/S sentences have distinct parses, so parse probes can measure the probability of each.

\paragraph{Experiment} We base our structural probes on \citeposs{eisape-etal-2022-probing} MLP action probes, as these are compatible with autoregressive models and incomplete inputs; most such probes are not. These probes take in the residual-stream representations of two words (from a fixed layer) and use a MLP to map them to one of three possible dependency relations: 1) the first word is a dependent of the second (\texttt{LEFT-ARC}); 2) vice-versa (\texttt{RIGHT-ARC}); or 3) no relation (\texttt{GEN}). Following \citet{eisape-etal-2022-probing}, we train probes to predict parser actions using parse-annotated data from the Penn TreeBank \citep{taylor2003penn}. As in \citet{eisape-etal-2022-probing}, our trained probes achieve high performance; see App.~\ref{app:structural-probe}.

With these probes, we evaluate our model's reading of ambiguous garden path sentences.\footnote{We verify that the structural probes' predictions on non-ambiguous sentences are sensible in App.~\ref{app:structural-probe}.} Crucial here is the dependency relation between each sentence's verb and final noun. The garden path reading of the sentence ``After the politician signed the bill'' would leave \textit{the bill} as a dependent (object) of \textit{signed}; in the non-GP case, there is no dependency relation. We record the probability of each relation,
averaged across all NP/Z and NP/S sentences.

\begin{figure}
    \centering
    \includegraphics[width=\linewidth]{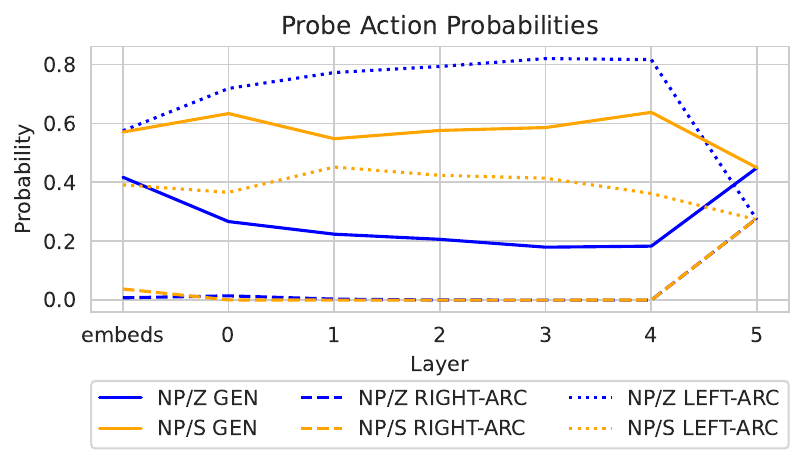}
    \caption{Mean probe action probability across layers. \texttt{GEN} corresponds to the non-GP reading, and \texttt{LEFT-ARC} to the GP reading (\texttt{RIGHT-ARC} is implausible). NP/Z sentences elicit primarily \texttt{LEFT-ARC}; NP/S elicits \texttt{GEN}. Both valid readings always receive non-zero probability.}
    \label{fig:probe-probs}
\end{figure}

\paragraph{Results} Our results (\Cref{fig:probe-probs}) show that while probes favor \texttt{LEFT-ARC} for NP/Z sentences, and \texttt{GEN} for NP/S, they assign moderate probability to both readings. This trend holds for all layers but the last, where probe performance is poor. This further supports the hypothesis that LMs consider both readings of garden-path sentences, as found in the feature analysis; in App.~\ref{app:probe-feature-consistency} we run AtP-IG on the probe and find the same features are responsible.

\section{Do LMs reanalyze, repair, or neither?}\label{sec:reanalysis-repair}
In humans, the semantics of garden-path readings can linger even after the sentence is complete \citep{christianson-2001-linger}: given ``The boy fed the chicken smiled.'', people often respond \emph{yes} to ``Did the boy feed the chicken?''. ``The boy'' is the recipient of \emph{fed}, but the garden-path reading suggests that it is the agent. 

What happens to LM representations after receiving such disambiguating information? If the LM relies on its original syntactic features, we might observe features at later token positions that adjudicate between different readings, analogous to a \textbf{repair}-based strategy. The later features could also upweight the correct reading independently of the original representations, analogous to \textbf{reanalysis}. 

\CR{We concretely operationalize these two hypotheses as follows. A model engages in repair if, at positions at or after the disambiguating token, it relies on previously-computed reading-specific syntactic features (e.g. subject or object detectors) to compute its output. The model may choose to select or ignore some of these previous features as part of the repair process. In contrast, a model engages in reanalysis if, at positions at or after the disambiguating token, it does not rely on previously-computed reading-specific syntactic features, and instead only relies on reading-agnostic features (e.g. word or part-of-speech detectors) to compute its output. The model might compute new reading-specific features at positions after the disambiguating token.}

\subsection{Behavioral Analysis}\label{subsec:readingcomp_behavioral}
We evaluate how models respond to garden path reading comprehension (GPRC) questions; past work suggests fine-tuned masked LMs exhibit lingering garden path effects \citep{irwin-etal-2023-bert}. Correct answers to GPRC questions indicate a correct (non-garden-path) reading of the sentence. For example, given ``The boy fed the chicken smiled.'', we ask ``Did the boy smile?'' (\emph{Yes}) or  ``Did the boy feed the chicken?''(\emph{No}). We craft \emph{Yes} and \emph{No} questions for each sentence, so models whose answers are random or constant will obtain 50\% accuracy.

\begin{table}
    \centering
    \begin{tabular}{lrrrrr}
        \toprule
        & \multicolumn{2}{c}{Yes/No QA} & \multicolumn{2}{c}{GPRC}\\
        \cmidrule(lr){2-3}\cmidrule(lr){4-5}
        Model & BoolQ & MCQA  & NPS & NPZ \\
        \midrule
        Pythia & 42.8 & 50.0 & 50.0 & 50.0 \\
        Gemma 2 & 70.7 & 90.0 & 83.3 & 70.9 \\
        \bottomrule
    \end{tabular}
    \caption{Accuracy on two QA datasets and garden-path reading comprehension (GPRC) questions (all zero-shot binary questions). Pythia performs poorly as it often outputs the same answer for all inputs, regardless of the question. Gemma 2 performs well on all tasks.}
    \label{tab:qa_results}
\end{table}

\paragraph{Experiment} We first verify whether our models can do question answering (QA) on less tricky binary QA datasets. Good performance here is a prerequisite for the following analyses to be valid. We evaluate on a binary version of multiple choice question answering (MCQA) from \citet{wiegreffe-etal-2024-answer}, where questions are of the form ``Question: Boxes are brown. What color are boxes?\texttt{\textbackslash{}n}A.\ green\texttt{\textbackslash{}n}B.\ brown{\textbackslash{}n}Answer:'', and the model must answer `` A'' or `` B''. We also evaluate on BoolQ \citep{clark-etal-2019-boolq}, a naturalistic QA dataset consisting of context passages followed by a yes/no question. For all tasks, we use a zero-shot setup: the model is prompted only with the question, context, and answer options. We measure accuracy as the frequency with which the model prefers the correct answer token to the incorrect one.

\paragraph{Results} Our results (\Cref{tab:qa_results}) indicate that only Gemma-2-2b performs well on all tasks;\footnote{We note that BoolQ is challenging: even otherwise well-performing models obtain close to 70--80\% performance, even with demonstrations. We thus take 70\% as positive evidence of the model's binary QA ability.} it also answers GPRC questions with above-chance accuracy, so we focus the rest of our analysis on it.

\subsection{Feature and Causal Analysis}
Ideally, a model answering GPRC questions should rely on features indicative of the input's parse. To verify this, we can measure the overlap between the features from \S\ref{subsec:feature} and those obtained via AtP-IG on the GPRC questions. We can also ablate the features from \S\ref{subsec:feature} and measure the performance of the model on GPRC questions.

\paragraph{Experiment} We discover sparse feature circuits for GPRC questions using AtP-IG, as in \S\ref{subsec:feature}. The prompts consist of complete sentences and questions, $m = p(\text{Yes}) - p(\text{No})$, and our score threshold is 0.05. We measure the overlap between the circuits from \S\ref{subsec:feature} (denoted $C_1$) and the GPRC circuits (denoted $C_2$) as the intersection-over-union (IoU) of $C_1$ and $C_2$'s features.

We also check if $C_1$'s features causally influence the GPRC task. As in \S\ref{subsec:causal}, we annotate $C_1$ features and place them in groups, like subject or object detectors (we causally verify these groups' relevance in App.~\ref{subsec:gemma-causal}). Then, we manipulate these features as in prior experiments, and record model accuracy: we upweight subject detectors and zero ablate object detectors to promote the non-garden-path reading, aiming to increase model accuracy; we do the reverse to decrease it.

\paragraph{Results} There is little feature overlap across circuits: the IoU is 0\% for NP/S, and 0.2\% for NP/Z. Accordingly, Gemma 2 does not rely extensively on features from $C_1$ to answer follow-up questions: performance changes little when intervening on these features. Indeed, the top GPRC features are unrelated to either parse; many are not syntax-sensitive, but are instead spurious features that promote \textit{Yes} or \textit{No}. \textit{Yes}-promoting features often activate on phrases related to agreement, such as ``Certainly'' or ``Of course''. Given that the effect of $C_1$'s syntax-sensitive features is not exactly 0, the model does use them to a non-zero extent; nonetheless, they explain little of the model's GPRC behavior.

This suggests that the Gemma 2 does not repair its previously constructed representations when answering follow-up questions about garden-path sentences; however, it does not appear to generate new syntactic features via reanalysis, either. While this behavior is more akin to reanalysis than repair, as features are not reused, we hypothesize that it reflects a process that is fundamentally different from that of reanalysis in humans. Namely, reanalysis in humans assumes that humans will construct new syntactic representations to answer follow-up questions; in contrast, although Gemma 2 constructs new features when answering follow-up questions, these features are not syntactic. Thus, while both models and humans rely on syntactic features when predicting the disambiguating word in garden path sentences, the same may not be true for predicting the answer to garden-path follow-up questions.

\section{General Discussion and Conclusions}
When conducting behavioral analyses, one must be cautious in (but not entirely averse to) imposing human-like cognitive abstractions onto LMs. Despite high performance on syntactic evaluations, we have observed that LMs rely on both human-like syntactic abstractions as well as spurious features. Indeed, many influential features activated on tokens before relevant syntactic information for \emph{either} reading of the sentence had appeared. This underscores the importance of mechanistic investigations of LM behaviors: even when models perform well, it may not always be for the reasons that an informed researcher would anticipate.

We have seen that LMs \emph{represent} ambiguities, holding on to multiple interpretations of partial sentences. However, it remains unclear if LMs deploy mechanisms that \emph{recognize} or adjudicate between mutually exclusive possibilities. The representation--recognition distinction is crucial: ambiguity has many functions (e.g., humor and politeness), but detecting these requires recognizing ambiguity as a meaningful signal. We leave the question of ambiguity recognition to future work.

LMs did not rely on prior features when answering garden path follow-up questions, indicating a lack of repair, but also did not generate any new syntactic features as might be expected via reanalysis. While we cannot definitively rule out the existence of syntactic reanalysis circuits, such features appear uninfluential in the GPRC circuits. We hope that future advances in sparse autoencoders and automated interpretability methods will enable us to more deeply understand sophisticated and large sparse feature circuits at scale.

\section*{Acknowledgments}
We thank Hadas Orgad for helpful feedback on an earlier version of this paper. M.H. is supported by an OpenAI Superalignment Fellowship. A.M. is supported by a postdoctoral fellowship under the Zuckerman STEM Leadership Program.

\section*{Limitations}
Our study has focused primarily on individual features. While we do make use of edges between features in our qualitative analysis, we have not causally verified what these edges signify. For example, are these AND or OR relations, NOT relations, or some more sophisticated type of feature combination? A deeper investigation could yield greater insights into how repair/reanalysis happens, and how past features remain relevant at later positions (or are made irrelevant).

We have analyzed two language models of significantly differing scales and slightly differing architectures/training setups. While we are confident in concluding that Transformer-based autoregressive language models are generally likely to encode the mechanisms we have discovered, the results could still be strengthened by extending the analysis to a models with more diverse training setups, scales, and architectures. It would be particularly interesting---and helpful in linking our results to the learnability literature---to observe whether these results hold for more cognitively plausible language models, such as those trained on more human-sized datasets \citep[e.g.,][]{warstadt-etal-2023-findings}.

\CR{While this study was motivated by the study of incremental sentence processing in LMs, we study only garden path sentences to facilitate answering RQ2 and RQ3. Further work could investigate incremental sentence processing in more typical partial sentences; this would clarify whether different mechanisms are used for unambiguous sentences.}

\section*{Author Contributions}
\begin{itemize}[noitemsep]
    \item Conceptualization: A.M., M.H.
    \item Experimentation:
    \begin{itemize}[noitemsep]
        \item Behavioral analysis: M.H.
        \item Feature circuit discovery: A.M., M.H.
        \item Feature annotation: M.H., A.M.
        \item Causal feature analyses (RQ1, RQ2): M.H.
        \item Structural probing: M.H.
        \item Reading comprehension questions: A.M.
        \item Causal feature analysis (RQ3): A.M.
    \end{itemize}
    \item Writing: M.H., A.M.
\end{itemize}

\bibliography{custom,anthology_0,anthology_1}

\begin{thebibliography}{76}
\providecommand{\natexlab}[1]{#1}

\bibitem[{Arehalli et~al.(2022)Arehalli, Dillon, and
  Linzen}]{arehalli-etal-2022-syntactic}
Suhas Arehalli, Brian Dillon, and Tal Linzen. 2022.
\newblock \href {https://doi.org/10.18653/v1/2022.conll-1.20} {Syntactic
  surprisal from neural models predicts, but underestimates, human processing
  difficulty from syntactic ambiguities}.
\newblock In \emph{Proceedings of the 26th Conference on Computational Natural
  Language Learning (CoNLL)}, pages 301--313, Abu Dhabi, United Arab Emirates
  (Hybrid). Association for Computational Linguistics.

\bibitem[{Arps et~al.(2022)Arps, Samih, Kallmeyer, and
  Sajjad}]{arps-etal-2022-probing}
David Arps, Younes Samih, Laura Kallmeyer, and Hassan Sajjad. 2022.
\newblock \href {https://doi.org/10.18653/v1/2022.findings-emnlp.502} {Probing
  for constituency structure in neural language models}.
\newblock In \emph{Findings of the Association for Computational Linguistics:
  EMNLP 2022}, pages 6738--6757, Abu Dhabi, United Arab Emirates. Association
  for Computational Linguistics.

\bibitem[{Biderman et~al.(2023)Biderman, Schoelkopf, Anthony, Bradley, O'Brien,
  Hallahan, Khan, Purohit, Prashanth, Raff, Skowron, Sutawika, and Van
  Der~Wal}]{biderman2023pythia}
Stella Biderman, Hailey Schoelkopf, Quentin Anthony, Herbie Bradley, Kyle
  O'Brien, Eric Hallahan, Mohammad~Aflah Khan, Shivanshu Purohit, USVSN~Sai
  Prashanth, Edward Raff, Aviya Skowron, Lintang Sutawika, and Oskar Van
  Der~Wal. 2023.
\newblock Pythia: a suite for analyzing large language models across training
  and scaling.
\newblock In \emph{Proceedings of the 40th International Conference on Machine
  Learning}, ICML'23. JMLR.org.

\bibitem[{Bills et~al.(2023)Bills, Cammarata, Mossing, Tillman, Gao, Goh,
  Sutskever, Leike, Wu, and Saunders}]{bills2023language}
Steven Bills, Nick Cammarata, Dan Mossing, Henk Tillman, Leo Gao, Gabriel Goh,
  Ilya Sutskever, Jan Leike, Jeff Wu, and William Saunders. 2023.
\newblock Language models can explain neurons in language models.
\newblock
  \url{https://openaipublic.blob.core.windows.net/neuron-explainer/paper/index.html}.

\bibitem[{Bolukbasi et~al.(2021)Bolukbasi, Pearce, Yuan, Coenen, Reif, Viégas,
  and Wattenberg}]{bolukbasi2021interpretability}
Tolga Bolukbasi, Adam Pearce, Ann Yuan, Andy Coenen, Emily Reif, Fernanda
  Viégas, and Martin Wattenberg. 2021.
\newblock \href {https://arxiv.org/abs/2104.07143} {An interpretability
  illusion for {BERT}}.
\newblock \emph{Preprint}, arXiv:2104.07143.

\bibitem[{Bricken et~al.(2023)Bricken, Templeton, Batson, Chen, Jermyn,
  Conerly, Turner, Anil, Denison, Askell, Lasenby, Wu, Kravec, Schiefer,
  Maxwell, Joseph, Hatfield-Dodds, Tamkin, Nguyen, McLean, Burke, Hume, Carter,
  Henighan, and Olah}]{bricken2023monosemanticity}
Trenton Bricken, Adly Templeton, Joshua Batson, Brian Chen, Adam Jermyn, Tom
  Conerly, Nick Turner, Cem Anil, Carson Denison, Amanda Askell, Robert
  Lasenby, Yifan Wu, Shauna Kravec, Nicholas Schiefer, Tim Maxwell, Nicholas
  Joseph, Zac Hatfield-Dodds, Alex Tamkin, Karina Nguyen, Brayden McLean,
  Josiah~E Burke, Tristan Hume, Shan Carter, Tom Henighan, and Christopher
  Olah. 2023.
\newblock \href
  {https://transformer-circuits.pub/2023/monosemantic-features/index.html}
  {Towards monosemanticity: Decomposing language models with dictionary
  learning}.
\newblock \emph{Transformer Circuits Thread}.

\bibitem[{Christianson et~al.(2001)Christianson, Hollingworth, Halliwell, and
  Ferreira}]{christianson-2001-linger}
Kiel Christianson, Andrew Hollingworth, John~F. Halliwell, and Fernanda
  Ferreira. 2001.
\newblock \href {https://doi.org/10.1006/cogp.2001.0752} {Thematic roles
  assigned along the garden path linger}.
\newblock \emph{Cognitive Psychology}, 42(4):368--407.

\bibitem[{Christianson et~al.(2006)Christianson, Williams, Zacks, and
  Ferreira}]{christianson2006younger}
Kiel Christianson, Carrick~C. Williams, Rose~T. Zacks, and Fernanda Ferreira.
  2006.
\newblock \href {https://doi.org/10.1207/s15326950dp4202\_6} {Younger and older
  adults' "good-enough" interpretations of garden-path sentences}.
\newblock \emph{Discourse Processes}, 42(2):205--238.
\newblock PMID: 17203135.

\bibitem[{Clark et~al.(2019{\natexlab{a}})Clark, Lee, Chang, Kwiatkowski,
  Collins, and Toutanova}]{clark-etal-2019-boolq}
Christopher Clark, Kenton Lee, Ming-Wei Chang, Tom Kwiatkowski, Michael
  Collins, and Kristina Toutanova. 2019{\natexlab{a}}.
\newblock \href {https://doi.org/10.18653/v1/N19-1300} {{B}ool{Q}: Exploring
  the surprising difficulty of natural yes/no questions}.
\newblock In \emph{Proceedings of the 2019 Conference of the North {A}merican
  Chapter of the Association for Computational Linguistics: Human Language
  Technologies, Volume 1 (Long and Short Papers)}, pages 2924--2936,
  Minneapolis, Minnesota. Association for Computational Linguistics.

\bibitem[{Clark et~al.(2019{\natexlab{b}})Clark, Khandelwal, Levy, and
  Manning}]{clark-etal-2019-bert}
Kevin Clark, Urvashi Khandelwal, Omer Levy, and Christopher~D. Manning.
  2019{\natexlab{b}}.
\newblock \href {https://doi.org/10.18653/v1/W19-4828} {What does {BERT} look
  at? an analysis of {BERT}{'}s attention}.
\newblock In \emph{Proceedings of the 2019 ACL Workshop BlackboxNLP: Analyzing
  and Interpreting Neural Networks for NLP}, pages 276--286, Florence, Italy.
  Association for Computational Linguistics.

\bibitem[{Conmy et~al.(2023)Conmy, Mavor-Parker, Lynch, Heimersheim, and
  Garriga-Alonso}]{conmy2023towards}
Arthur Conmy, Augustine Mavor-Parker, Aengus Lynch, Stefan Heimersheim, and
  Adri\`{a} Garriga-Alonso. 2023.
\newblock \href
  {https://proceedings.neurips.cc/paper_files/paper/2023/file/34e1dbe95d34d7ebaf99b9bcaeb5b2be-Paper-Conference.pdf}
  {Towards automated circuit discovery for mechanistic interpretability}.
\newblock In \emph{Advances in Neural Information Processing Systems},
  volume~36, pages 16318--16352. Curran Associates, Inc.

\bibitem[{Eisape et~al.(2022)Eisape, Gangireddy, Levy, and
  Kim}]{eisape-etal-2022-probing}
Tiwalayo Eisape, Vineet Gangireddy, Roger Levy, and Yoon Kim. 2022.
\newblock \href {https://doi.org/10.18653/v1/2022.findings-emnlp.203} {Probing
  for incremental parse states in autoregressive language models}.
\newblock In \emph{Findings of the Association for Computational Linguistics:
  EMNLP 2022}, pages 2801--2813, Abu Dhabi, United Arab Emirates. Association
  for Computational Linguistics.

\bibitem[{Elazar et~al.(2021)Elazar, Ravfogel, Jacovi, and
  Goldberg}]{elazar-etal-2021-amnesic}
Yanai Elazar, Shauli Ravfogel, Alon Jacovi, and Yoav Goldberg. 2021.
\newblock \href {https://doi.org/10.1162/tacl_a_00359} {Amnesic probing:
  Behavioral explanation with amnesic counterfactuals}.
\newblock \emph{Transactions of the Association for Computational Linguistics},
  9:160--175.

\bibitem[{Elhage et~al.(2022)Elhage, Hume, Olsson, Schiefer, Henighan, Kravec,
  Hatfield-Dodds, Lasenby, Drain, Chen, Grosse, McCandlish, Kaplan, Amodei,
  Wattenberg, and Olah}]{elhage2022toymodelssuperposition}
Nelson Elhage, Tristan Hume, Catherine Olsson, Nicholas Schiefer, Tom Henighan,
  Shauna Kravec, Zac Hatfield-Dodds, Robert Lasenby, Dawn Drain, Carol Chen,
  Roger Grosse, Sam McCandlish, Jared Kaplan, Dario Amodei, Martin Wattenberg,
  and Christopher Olah. 2022.
\newblock \href {https://arxiv.org/abs/2209.10652} {Toy models of
  superposition}.
\newblock \emph{Preprint}, arXiv:2209.10652.

\bibitem[{Finlayson et~al.(2021)Finlayson, Mueller, Gehrmann, Shieber, Linzen,
  and Belinkov}]{finlayson-etal-2021-causal}
Matthew Finlayson, Aaron Mueller, Sebastian Gehrmann, Stuart Shieber, Tal
  Linzen, and Yonatan Belinkov. 2021.
\newblock \href {https://doi.org/10.18653/v1/2021.acl-long.144} {Causal
  analysis of syntactic agreement mechanisms in neural language models}.
\newblock In \emph{Proceedings of the 59th Annual Meeting of the Association
  for Computational Linguistics and the 11th International Joint Conference on
  Natural Language Processing (Volume 1: Long Papers)}, pages 1828--1843,
  Online. Association for Computational Linguistics.

\bibitem[{Fiotto-Kaufman et~al.(2024)Fiotto-Kaufman, Loftus, Todd, Brinkmann,
  Juang, Pal, Rager, Mueller, Marks, Sharma, Lucchetti, Ripa, Belfki, Prakash,
  Multani, Brodley, Guha, Bell, Wallace, and Bau}]{fiottokaufman2024nnsight}
Jaden Fiotto-Kaufman, Alexander~R Loftus, Eric Todd, Jannik Brinkmann, Caden
  Juang, Koyena Pal, Can Rager, Aaron Mueller, Samuel Marks, Arnab~Sen Sharma,
  Francesca Lucchetti, Michael Ripa, Adam Belfki, Nikhil Prakash, Sumeet
  Multani, Carla Brodley, Arjun Guha, Jonathan Bell, Byron Wallace, and David
  Bau. 2024.
\newblock \href {https://arxiv.org/abs/2407.14561} {{NNsight} and {NDIF}:
  Democratizing access to foundation model internals}.
\newblock \emph{Preprint}, arXiv:2407.14561.

\bibitem[{Fodor et~al.(1974)Fodor, Bever, and Garrett}]{fodor1974psychology}
J.A. Fodor, T.G. Bever, and M.F. Garrett. 1974.
\newblock \emph{The Psychology of Language: An Introduction to
  Psycholinguistics and Generative Grammar}.
\newblock McGraw-Hill Series on Computer Communications. McGraw-Hill.

\bibitem[{Fodor and Ferreira(1998)}]{fodor1998reanalysis}
Janet Fodor and Fernanda Ferreira, editors. 1998.
\newblock \emph{Reanalysis in sentence processing}, volume~21.
\newblock Springer Science \& Business Media.

\bibitem[{Frazier(1979)}]{frazier1979comprehending}
Lyn Frazier. 1979.
\newblock \emph{On Comprehending Sentences: Syntactic Parsing Strategies}.
\newblock Ph.D. thesis, University of Connecticut.

\bibitem[{Frazier(1987)}]{frazier1987sentence}
Lyn Frazier. 1987.
\newblock Sentence processing: A tutorial review.
\newblock \emph{Attention and performance XII}, pages 559--586.

\bibitem[{Gao et~al.(2020)Gao, Biderman, Black, Golding, Hoppe, Foster, Phang,
  He, Thite, Nabeshima, Presser, and Leahy}]{gao2020pile}
Leo Gao, Stella Biderman, Sid Black, Laurence Golding, Travis Hoppe, Charles
  Foster, Jason Phang, Horace He, Anish Thite, Noa Nabeshima, Shawn Presser,
  and Connor Leahy. 2020.
\newblock \href {https://arxiv.org/abs/2101.00027} {The pile: An 800gb dataset
  of diverse text for language modeling}.
\newblock \emph{Preprint}, arXiv:2101.00027.

\bibitem[{{Gemma Team} et~al.(2024){Gemma Team}, Riviere, Pathak, Sessa,
  Hardin, Bhupatiraju, Hussenot, Mesnard, Shahriari, Ramé, Ferret, Liu, Tafti,
  Friesen, Casbon, Ramos, Kumar, Lan, Jerome, Tsitsulin, Vieillard, Stanczyk,
  Girgin, Momchev, Hoffman, Thakoor, Grill, Neyshabur, Bachem, Walton, Severyn,
  Parrish, Ahmad, Hutchison, Abdagic, Carl, Shen, Brock, Coenen, Laforge,
  Paterson, Bastian, Piot, Wu, Royal, Chen, Kumar, Perry, Welty,
  Choquette-Choo, Sinopalnikov, Weinberger, Vijaykumar, Rogozińska, Herbison,
  Bandy, Wang, Noland, Moreira, Senter, Eltyshev, Visin, Rasskin, Wei, Cameron,
  Martins, Hashemi, Klimczak-Plucińska, Batra, Dhand, Nardini, Mein, Zhou,
  Svensson, Stanway, Chan, Zhou, Carrasqueira, Iljazi, Becker, Fernandez, van
  Amersfoort, Gordon, Lipschultz, Newlan, yeong Ji, Mohamed, Badola, Black,
  Millican, McDonell, Nguyen, Sodhia, Greene, Sjoesund, Usui, Sifre, Heuermann,
  Lago, McNealus, Soares, Kilpatrick, Dixon, Martins, Reid, Singh, Iverson,
  Görner, Velloso, Wirth, Davidow, Miller, Rahtz, Watson, Risdal, Kazemi,
  Moynihan, Zhang, Kahng, Park, Rahman, Khatwani, Dao, Bardoliwalla,
  Devanathan, Dumai, Chauhan, Wahltinez, Botarda, Barnes, Barham, Michel, Jin,
  Georgiev, Culliton, Kuppala, Comanescu, Merhej, Jana, Rokni, Agarwal,
  Mullins, Saadat, Carthy, Perrin, Arnold, Krause, Dai, Garg, Sheth, Ronstrom,
  Chan, Jordan, Yu, Eccles, Hennigan, Kocisky, Doshi, Jain, Yadav, Meshram,
  Dharmadhikari, Barkley, Wei, Ye, Han, Kwon, Xu, Shen, Gong, Wei, Cotruta,
  Kirk, Rao, Giang, Peran, Warkentin, Collins, Barral, Ghahramani, Hadsell,
  Sculley, Banks, Dragan, Petrov, Vinyals, Dean, Hassabis, Kavukcuoglu,
  Farabet, Buchatskaya, Borgeaud, Fiedel, Joulin, Kenealy, Dadashi, and
  Andreev}]{gemmateam2024gemma2improvingopen}
{Gemma Team}, Morgane Riviere, Shreya Pathak, Pier~Giuseppe Sessa, Cassidy
  Hardin, Surya Bhupatiraju, Léonard Hussenot, Thomas Mesnard, Bobak
  Shahriari, Alexandre Ramé, Johan Ferret, Peter Liu, Pouya Tafti, Abe
  Friesen, Michelle Casbon, Sabela Ramos, Ravin Kumar, Charline~Le Lan, Sammy
  Jerome, Anton Tsitsulin, Nino Vieillard, Piotr Stanczyk, Sertan Girgin,
  Nikola Momchev, Matt Hoffman, Shantanu Thakoor, Jean-Bastien Grill, Behnam
  Neyshabur, Olivier Bachem, Alanna Walton, Aliaksei Severyn, Alicia Parrish,
  Aliya Ahmad, Allen Hutchison, Alvin Abdagic, Amanda Carl, Amy Shen, Andy
  Brock, Andy Coenen, Anthony Laforge, Antonia Paterson, Ben Bastian, Bilal
  Piot, Bo~Wu, Brandon Royal, Charlie Chen, Chintu Kumar, Chris Perry, Chris
  Welty, Christopher~A. Choquette-Choo, Danila Sinopalnikov, David Weinberger,
  Dimple Vijaykumar, Dominika Rogozińska, Dustin Herbison, Elisa Bandy, Emma
  Wang, Eric Noland, Erica Moreira, Evan Senter, Evgenii Eltyshev, Francesco
  Visin, Gabriel Rasskin, Gary Wei, Glenn Cameron, Gus Martins, Hadi Hashemi,
  Hanna Klimczak-Plucińska, Harleen Batra, Harsh Dhand, Ivan Nardini, Jacinda
  Mein, Jack Zhou, James Svensson, Jeff Stanway, Jetha Chan, Jin~Peng Zhou,
  Joana Carrasqueira, Joana Iljazi, Jocelyn Becker, Joe Fernandez, Joost van
  Amersfoort, Josh Gordon, Josh Lipschultz, Josh Newlan, Ju~yeong Ji, Kareem
  Mohamed, Kartikeya Badola, Kat Black, Katie Millican, Keelin McDonell, Kelvin
  Nguyen, Kiranbir Sodhia, Kish Greene, Lars~Lowe Sjoesund, Lauren Usui,
  Laurent Sifre, Lena Heuermann, Leticia Lago, Lilly McNealus, Livio~Baldini
  Soares, Logan Kilpatrick, Lucas Dixon, Luciano Martins, Machel Reid,
  Manvinder Singh, Mark Iverson, Martin Görner, Mat Velloso, Mateo Wirth, Matt
  Davidow, Matt Miller, Matthew Rahtz, Matthew Watson, Meg Risdal, Mehran
  Kazemi, Michael Moynihan, Ming Zhang, Minsuk Kahng, Minwoo Park, Mofi Rahman,
  Mohit Khatwani, Natalie Dao, Nenshad Bardoliwalla, Nesh Devanathan, Neta
  Dumai, Nilay Chauhan, Oscar Wahltinez, Pankil Botarda, Parker Barnes, Paul
  Barham, Paul Michel, Pengchong Jin, Petko Georgiev, Phil Culliton, Pradeep
  Kuppala, Ramona Comanescu, Ramona Merhej, Reena Jana, Reza~Ardeshir Rokni,
  Rishabh Agarwal, Ryan Mullins, Samaneh Saadat, Sara~Mc Carthy, Sarah Perrin,
  Sébastien M.~R. Arnold, Sebastian Krause, Shengyang Dai, Shruti Garg, Shruti
  Sheth, Sue Ronstrom, Susan Chan, Timothy Jordan, Ting Yu, Tom Eccles, Tom
  Hennigan, Tomas Kocisky, Tulsee Doshi, Vihan Jain, Vikas Yadav, Vilobh
  Meshram, Vishal Dharmadhikari, Warren Barkley, Wei Wei, Wenming Ye, Woohyun
  Han, Woosuk Kwon, Xiang Xu, Zhe Shen, Zhitao Gong, Zichuan Wei, Victor
  Cotruta, Phoebe Kirk, Anand Rao, Minh Giang, Ludovic Peran, Tris Warkentin,
  Eli Collins, Joelle Barral, Zoubin Ghahramani, Raia Hadsell, D.~Sculley,
  Jeanine Banks, Anca Dragan, Slav Petrov, Oriol Vinyals, Jeff Dean, Demis
  Hassabis, Koray Kavukcuoglu, Clement Farabet, Elena Buchatskaya, Sebastian
  Borgeaud, Noah Fiedel, Armand Joulin, Kathleen Kenealy, Robert Dadashi, and
  Alek Andreev. 2024.
\newblock \href {https://arxiv.org/abs/2408.00118} {Gemma 2: Improving open
  language models at a practical size}.
\newblock \emph{Preprint}, arXiv:2408.00118.

\bibitem[{Gibson and Pearlmutter(2000)}]{gibson2000distinguishing}
Edward Gibson and Neal~J Pearlmutter. 2000.
\newblock Distinguishing serial and parallel parsing.
\newblock \emph{Journal of Psycholinguistic Research}, 29:231--240.

\bibitem[{Gibson(1991)}]{gibson1991computational}
Edward Albert~Fletcher Gibson. 1991.
\newblock \emph{A computational theory of human linguistic processing: memory
  limitations and processing breakdown}.
\newblock Ph.D. thesis, Carnegie Mellon University, USA.
\newblock UMI Order No. GAX91-26944.

\bibitem[{Goldberg(2019)}]{goldberg2019assessingbertssyntacticabilities}
Yoav Goldberg. 2019.
\newblock \href {https://arxiv.org/abs/1901.05287} {Assessing {BERT}'s
  syntactic abilities}.
\newblock \emph{Preprint}, arXiv:1901.05287.

\bibitem[{Gorrell(1987)}]{gorrell1987studies}
Paul~Griffith Gorrell. 1987.
\newblock \emph{Studies of human syntactic processing: Ranked-parallel versus
  serial models}.
\newblock Ph.D. thesis, University of Connecticut.

\bibitem[{Grodner et~al.(2003)Grodner, Gibson, Argaman, and
  Babyonyshev}]{grodner2003against}
Daniel Grodner, Edward Gibson, Vered Argaman, and Maria Babyonyshev. 2003.
\newblock Against repair-based reanalysis in sentence comprehension.
\newblock \emph{Journal of psycholinguistic research}, 32(2):141--166.

\bibitem[{Hanna et~al.(2023)Hanna, Liu, and Variengien}]{hanna2023gpt2}
Michael Hanna, Ollie Liu, and Alexandre Variengien. 2023.
\newblock \href
  {https://proceedings.neurips.cc/paper_files/paper/2023/file/efbba7719cc5172d175240f24be11280-Paper-Conference.pdf}
  {How does {GPT}-2 compute greater-than?: Interpreting mathematical abilities
  in a pre-trained language model}.
\newblock In \emph{Advances in Neural Information Processing Systems},
  volume~36, pages 76033--76060. Curran Associates, Inc.

\bibitem[{Hanna et~al.(2024)Hanna, Pezzelle, and Belinkov}]{hanna2024have}
Michael Hanna, Sandro Pezzelle, and Yonatan Belinkov. 2024.
\newblock \href {https://openreview.net/forum?id=TZ0CCGDcuT} {Have faith in
  faithfulness: Going beyond circuit overlap when finding model mechanisms}.
\newblock In \emph{First Conference on Language Modeling}.

\bibitem[{Hewitt and Manning(2019)}]{hewitt-manning-2019-structural}
John Hewitt and Christopher~D. Manning. 2019.
\newblock \href {https://doi.org/10.18653/v1/N19-1419} {{A} structural probe
  for finding syntax in word representations}.
\newblock In \emph{Proceedings of the 2019 Conference of the North {A}merican
  Chapter of the Association for Computational Linguistics: Human Language
  Technologies, Volume 1 (Long and Short Papers)}, pages 4129--4138,
  Minneapolis, Minnesota. Association for Computational Linguistics.

\bibitem[{Hinton et~al.(1986)Hinton, McClelland, and
  Rumelhart}]{hinton1986distributed}
G.~E. Hinton, J.~L. McClelland, and D.~E. Rumelhart. 1986.
\newblock \emph{Distributed representations}, page 77–109.
\newblock MIT Press, Cambridge, MA, USA.

\bibitem[{Htut et~al.(2019)Htut, Phang, Bordia, and Bowman}]{htut2019attention}
Phu~Mon Htut, Jason Phang, Shikha Bordia, and Samuel~R. Bowman. 2019.
\newblock \href {https://arxiv.org/abs/1911.12246} {Do attention heads in bert
  track syntactic dependencies?}
\newblock \emph{Preprint}, arXiv:1911.12246.

\bibitem[{Hu and Frank(2024)}]{hu2024auxiliary}
Jennifer Hu and Michael Frank. 2024.
\newblock \href {https://openreview.net/forum?id=U5BUzSn4tD} {Auxiliary task
  demands mask the capabilities of smaller language models}.
\newblock In \emph{First Conference on Language Modeling}.

\bibitem[{Hu et~al.(2020)Hu, Gauthier, Qian, Wilcox, and
  Levy}]{hu-etal-2020-systematic}
Jennifer Hu, Jon Gauthier, Peng Qian, Ethan Wilcox, and Roger Levy. 2020.
\newblock \href {https://doi.org/10.18653/v1/2020.acl-main.158} {A systematic
  assessment of syntactic generalization in neural language models}.
\newblock In \emph{Proceedings of the 58th Annual Meeting of the Association
  for Computational Linguistics}, pages 1725--1744, Online. Association for
  Computational Linguistics.

\bibitem[{Hu and Levy(2023)}]{hu-levy-2023-prompting}
Jennifer Hu and Roger Levy. 2023.
\newblock \href {https://doi.org/10.18653/v1/2023.emnlp-main.306} {Prompting is
  not a substitute for probability measurements in large language models}.
\newblock In \emph{Proceedings of the 2023 Conference on Empirical Methods in
  Natural Language Processing}, pages 5040--5060, Singapore. Association for
  Computational Linguistics.

\bibitem[{Huang et~al.(2023)Huang, Geiger, D{'}Oosterlinck, Wu, and
  Potts}]{huang-etal-2023-rigorously}
Jing Huang, Atticus Geiger, Karel D{'}Oosterlinck, Zhengxuan Wu, and
  Christopher Potts. 2023.
\newblock \href {https://doi.org/10.18653/v1/2023.blackboxnlp-1.24} {Rigorously
  assessing natural language explanations of neurons}.
\newblock In \emph{Proceedings of the 6th BlackboxNLP Workshop: Analyzing and
  Interpreting Neural Networks for NLP}, pages 317--331, Singapore. Association
  for Computational Linguistics.

\bibitem[{Huang et~al.(2024)Huang, Arehalli, Kugemoto, Muxica, Prasad, Dillon,
  and Linzen}]{huang2024large}
Kuan-Jung Huang, Suhas Arehalli, Mari Kugemoto, Christian Muxica, Grusha
  Prasad, Brian Dillon, and Tal Linzen. 2024.
\newblock \href {https://doi.org/10.1016/j.jml.2024.104510} {Large-scale
  benchmark yields no evidence that language model surprisal explains syntactic
  disambiguation difficulty}.
\newblock \emph{Journal of Memory and Language}, 137:104510.

\bibitem[{Irwin et~al.(2023)Irwin, Wilson, and Marantz}]{irwin-etal-2023-bert}
Tovah Irwin, Kyra Wilson, and Alec Marantz. 2023.
\newblock \href {https://doi.org/10.18653/v1/2023.eacl-main.235} {{BERT} shows
  garden path effects}.
\newblock In \emph{Proceedings of the 17th Conference of the European Chapter
  of the Association for Computational Linguistics}, pages 3220--3232,
  Dubrovnik, Croatia. Association for Computational Linguistics.

\bibitem[{Jawahar et~al.(2019)Jawahar, Sagot, and
  Seddah}]{jawahar-etal-2019-bert}
Ganesh Jawahar, Beno{\^\i}t Sagot, and Djam{\'e} Seddah. 2019.
\newblock \href {https://doi.org/10.18653/v1/P19-1356} {What does {BERT} learn
  about the structure of language?}
\newblock In \emph{Proceedings of the 57th Annual Meeting of the Association
  for Computational Linguistics}, pages 3651--3657, Florence, Italy.
  Association for Computational Linguistics.

\bibitem[{Jurafsky(1996)}]{jurafsky1996probabilistic}
Daniel Jurafsky. 1996.
\newblock \href {https://doi.org/10.1207/s15516709cog2002\_1} {A probabilistic
  model of lexical and syntactic access and disambiguation}.
\newblock \emph{Cognitive Science}, 20(2):137--194.

\bibitem[{Lasri et~al.(2022)Lasri, Pimentel, Lenci, Poibeau, and
  Cotterell}]{lasri-etal-2022-probing}
Karim Lasri, Tiago Pimentel, Alessandro Lenci, Thierry Poibeau, and Ryan
  Cotterell. 2022.
\newblock \href {https://doi.org/10.18653/v1/2022.acl-long.603} {Probing for
  the usage of grammatical number}.
\newblock In \emph{Proceedings of the 60th Annual Meeting of the Association
  for Computational Linguistics (Volume 1: Long Papers)}, pages 8818--8831,
  Dublin, Ireland. Association for Computational Linguistics.

\bibitem[{Lewis(1998)}]{lewis1998reanalysis}
Richard~L. Lewis. 1998.
\newblock \href {https://doi.org/10.1007/978-94-015-9070-9_8} {\emph{Reanalysis
  and Limited Repair Parsing: Leaping off the Garden Path}}, chapter Reanalysis
  and Limited Repair Parsing: Leaping off the Garden Path.
\newblock Springer Netherlands, Dordrecht.

\bibitem[{Lewis(2000)}]{lewis2000falsifying}
Richard~L Lewis. 2000.
\newblock Falsifying serial and parallel parsing models: Empirical conundrums
  and an overlooked paradigm.
\newblock \emph{Journal of Psycholinguistic Research}, 29:241--248.

\bibitem[{Li et~al.(2024)Li, Feng, Narang, Peng, Cai, Shah, and
  Varma}]{li2024incremental}
Andrew Li, Xianle Feng, Siddhant Narang, Austin Peng, Tianle Cai, Raj~Sanjay
  Shah, and Sashank Varma. 2024.
\newblock Incremental comprehension of garden-path sentences by large language
  models: Semantic interpretation, syntactic re-analysis, and attention.
\newblock In \emph{Proceedings of the Annual Meeting of the Cognitive Science
  Society}, volume~46.

\bibitem[{Lieberum et~al.(2024)Lieberum, Rajamanoharan, Conmy, Smith, Sonnerat,
  Varma, Kramar, Dragan, Shah, and Nanda}]{lieberum-etal-2024-gemma}
Tom Lieberum, Senthooran Rajamanoharan, Arthur Conmy, Lewis Smith, Nicolas
  Sonnerat, Vikrant Varma, Janos Kramar, Anca Dragan, Rohin Shah, and Neel
  Nanda. 2024.
\newblock \href {https://doi.org/10.18653/v1/2024.blackboxnlp-1.19} {Gemma
  scope: Open sparse autoencoders everywhere all at once on gemma 2}.
\newblock In \emph{Proceedings of the 7th BlackboxNLP Workshop: Analyzing and
  Interpreting Neural Networks for NLP}, pages 278--300, Miami, Florida, US.
  Association for Computational Linguistics.

\bibitem[{Lin and Bloom(2023)}]{neuronpedia}
Johnny Lin and Joseph Bloom. 2023.
\newblock \href {https://www.neuronpedia.org} {Neuronpedia: Interactive
  reference and tooling for analyzing neural networks}.
\newblock Software available from neuronpedia.org.

\bibitem[{Marks et~al.(2024)Marks, Rager, Michaud, Belinkov, Bau, and
  Mueller}]{marks2024sparse}
Samuel Marks, Can Rager, Eric~J. Michaud, Yonatan Belinkov, David Bau, and
  Aaron Mueller. 2024.
\newblock \href {https://arxiv.org/abs/2403.19647} {Sparse feature circuits:
  Discovering and editing interpretable causal graphs in language models}.
\newblock \emph{Preprint}, arXiv:2403.19647.

\bibitem[{Marslen-Wilson(1975)}]{marslenwilson1975sentence}
William~D. Marslen-Wilson. 1975.
\newblock \href {https://doi.org/10.1126/science.189.4198.226} {Sentence
  perception as an interactive parallel process}.
\newblock \emph{Science}, 189(4198):226--228.

\bibitem[{Maudslay et~al.(2020)Maudslay, Valvoda, Pimentel, Williams, and
  Cotterell}]{hall-maudslay-etal-2020-tale}
Rowan~Hall Maudslay, Josef Valvoda, Tiago Pimentel, Adina Williams, and Ryan
  Cotterell. 2020.
\newblock \href {https://doi.org/10.18653/v1/2020.acl-main.659} {A tale of a
  probe and a parser}.
\newblock In \emph{Proceedings of the 58th Annual Meeting of the Association
  for Computational Linguistics}, pages 7389--7395, Online. Association for
  Computational Linguistics.

\bibitem[{Miller et~al.(2024)Miller, Chughtai, and
  Saunders}]{miller2024transformer}
Joseph Miller, Bilal Chughtai, and William Saunders. 2024.
\newblock \href {https://openreview.net/forum?id=zSf8PJyQb2} {Transformer
  circuit evaluation metrics are not robust}.
\newblock In \emph{First Conference on Language Modeling}.

\bibitem[{Nanda(2023)}]{nanda2023attribution}
Neel Nanda. 2023.
\newblock \href
  {https://www.neelnanda.io/mechanistic-interpretability/attribution-patching}
  {Attribution {Patching}: {Activation} {Patching} {At} {Industrial} {Scale}}.

\bibitem[{Newman et~al.(2021)Newman, Ang, Gong, and
  Hewitt}]{newman-etal-2021-refining}
Benjamin Newman, Kai-Siang Ang, Julia Gong, and John Hewitt. 2021.
\newblock \href {https://doi.org/10.18653/v1/2021.naacl-main.290} {Refining
  targeted syntactic evaluation of language models}.
\newblock In \emph{Proceedings of the 2021 Conference of the North American
  Chapter of the Association for Computational Linguistics: Human Language
  Technologies}, pages 3710--3723, Online. Association for Computational
  Linguistics.

\bibitem[{Nivre(2004)}]{nivre2004incrementality}
Joakim Nivre. 2004.
\newblock Incrementality in deterministic dependency parsing.
\newblock In \emph{Proceedings of the workshop on incremental parsing: Bringing
  engineering and cognition together}, pages 50--57.

\bibitem[{Oh and Schuler(2023)}]{oh-schuler-2023-surprisal}
Byung-Doh Oh and William Schuler. 2023.
\newblock \href {https://doi.org/10.1162/tacl_a_00548} {Why does surprisal from
  larger transformer-based language models provide a poorer fit to human
  reading times?}
\newblock \emph{Transactions of the Association for Computational Linguistics},
  11:336--350.

\bibitem[{Olah et~al.(2020)Olah, Cammarata, Schubert, Goh, Petrov, and
  Carter}]{olah2020zoom}
Chris Olah, Nick Cammarata, Ludwig Schubert, Gabriel Goh, Michael Petrov, and
  Shan Carter. 2020.
\newblock \href {https://doi.org/10.23915/distill.00024.001} {Zoom in: An
  introduction to circuits}.
\newblock \emph{Distill}.
\newblock Https://distill.pub/2020/circuits/zoom-in.

\bibitem[{Olah et~al.(2017)Olah, Mordvintsev, and Schubert}]{olah2017feature}
Chris Olah, Alexander Mordvintsev, and Ludwig Schubert. 2017.
\newblock \href {https://doi.org/10.23915/distill.00007} {Feature
  visualization}.
\newblock \emph{Distill}.

\bibitem[{Paszke et~al.(2019)Paszke, Gross, Massa, Lerer, Bradbury, Chanan,
  Killeen, Lin, Gimelshein, Antiga, Desmaison, Köpf, Yang, DeVito, Raison,
  Tejani, Chilamkurthy, Steiner, Fang, Bai, and Chintala}]{paszke2019pytorch}
Adam Paszke, Sam Gross, Francisco Massa, Adam Lerer, James Bradbury, Gregory
  Chanan, Trevor Killeen, Zeming Lin, Natalia Gimelshein, Luca Antiga, Alban
  Desmaison, Andreas Köpf, Edward Yang, Zach DeVito, Martin Raison, Alykhan
  Tejani, Sasank Chilamkurthy, Benoit Steiner, Lu~Fang, Junjie Bai, and Soumith
  Chintala. 2019.
\newblock \href {https://arxiv.org/abs/1912.01703} {Pytorch: An imperative
  style, high-performance deep learning library}.
\newblock \emph{Preprint}, arXiv:1912.01703.

\bibitem[{Pearl(2001)}]{pearl2001indirect}
Judea Pearl. 2001.
\newblock Direct and indirect effects.
\newblock In \emph{Proceedings of the Seventeenth Conference on Uncertainty in
  Artificial Intelligence}, UAI'01, page 411–420, San Francisco, CA, USA.
  Morgan Kaufmann Publishers Inc.

\bibitem[{Ravichander et~al.(2021)Ravichander, Belinkov, and
  Hovy}]{ravichander-etal-2021-probing}
Abhilasha Ravichander, Yonatan Belinkov, and Eduard Hovy. 2021.
\newblock \href {https://doi.org/10.18653/v1/2021.eacl-main.295} {Probing the
  probing paradigm: Does probing accuracy entail task relevance?}
\newblock In \emph{Proceedings of the 16th Conference of the European Chapter
  of the Association for Computational Linguistics: Main Volume}, pages
  3363--3377, Online. Association for Computational Linguistics.

\bibitem[{Sajjad et~al.(2022)Sajjad, Durrani, and
  Dalvi}]{sajjad-etal-2022-neuron}
Hassan Sajjad, Nadir Durrani, and Fahim Dalvi. 2022.
\newblock \href {https://doi.org/10.1162/tacl_a_00519} {Neuron-level
  interpretation of deep {NLP} models: A survey}.
\newblock \emph{Transactions of the Association for Computational Linguistics},
  10:1285--1303.

\bibitem[{Smolensky(1986)}]{smolensky1986distributed}
Paul Smolensky. 1986.
\newblock \emph{Neural and conceptual interpretation of PDP models}, page
  390–431.
\newblock MIT Press, Cambridge, MA, USA.

\bibitem[{Sturt et~al.(1999)Sturt, Pickering, and
  Crocker}]{sturt1999structural}
Patrick Sturt, Martin~J. Pickering, and Matthew~W. Crocker. 1999.
\newblock \href {https://doi.org/10.1006/jmla.1998.2606} {Structural change and
  reanalysis difficulty in language comprehension}.
\newblock \emph{Journal of Memory and Language}, 40(1):136--150.

\bibitem[{Sundararajan et~al.(2017)Sundararajan, Taly, and
  Yan}]{sundarajan201axiomatic}
Mukund Sundararajan, Ankur Taly, and Qiqi Yan. 2017.
\newblock Axiomatic attribution for deep networks.
\newblock In \emph{Proceedings of the 34th International Conference on Machine
  Learning - Volume 70}, ICML'17, page 3319–3328. JMLR.org.

\bibitem[{Taylor et~al.(2003)Taylor, Marcus, and Santorini}]{taylor2003penn}
Ann Taylor, Mitchell Marcus, and Beatrice Santorini. 2003.
\newblock The penn treebank: an overview.
\newblock \emph{Treebanks: Building and using parsed corpora}, pages 5--22.

\bibitem[{Tenney et~al.(2019)Tenney, Xia, Chen, Wang, Poliak, McCoy, Kim,
  Durme, Bowman, Das, and Pavlick}]{tenney2018what}
Ian Tenney, Patrick Xia, Berlin Chen, Alex Wang, Adam Poliak, R~Thomas McCoy,
  Najoung Kim, Benjamin~Van Durme, Sam Bowman, Dipanjan Das, and Ellie Pavlick.
  2019.
\newblock \href {https://openreview.net/forum?id=SJzSgnRcKX} {What do you learn
  from context? probing for sentence structure in contextualized word
  representations}.
\newblock In \emph{International Conference on Learning Representations}.

\bibitem[{van Gompel and Pickering(2007)}]{vanGompel2007syntactic}
Roger P.~G. van Gompel and Martin~J. Pickering. 2007.
\newblock \href {https://doi.org/10.1093/oxfordhb/9780198568971.013.0017}
  {{Syntactic parsing}}.
\newblock In \emph{{The Oxford Handbook of Psycholinguistics}}. Oxford
  University Press.

\bibitem[{Van~Schijndel and Linzen(2018)}]{van2018modeling}
Marten Van~Schijndel and Tal Linzen. 2018.
\newblock Modeling garden path effects without explicit hierarchical syntax.
\newblock In \emph{Proceedings of the 40th Annual Meeting of the Cognitive
  Science Society (CogSci 2018)t}.

\bibitem[{Van~Schijndel and Linzen(2021)}]{van2021single}
Marten Van~Schijndel and Tal Linzen. 2021.
\newblock Single-stage prediction models do not explain the magnitude of
  syntactic disambiguation difficulty.
\newblock \emph{Cognitive science}, 45(6):e12988.

\bibitem[{Vig and Belinkov(2019)}]{vig-belinkov-2019-analyzing}
Jesse Vig and Yonatan Belinkov. 2019.
\newblock \href {https://doi.org/10.18653/v1/W19-4808} {Analyzing the structure
  of attention in a transformer language model}.
\newblock In \emph{Proceedings of the 2019 ACL Workshop BlackboxNLP: Analyzing
  and Interpreting Neural Networks for NLP}, pages 63--76, Florence, Italy.
  Association for Computational Linguistics.

\bibitem[{Vig et~al.(2020)Vig, Gehrmann, Belinkov, Qian, Nevo, Singer, and
  Shieber}]{vig2020investigating}
Jesse Vig, Sebastian Gehrmann, Yonatan Belinkov, Sharon Qian, Daniel Nevo,
  Yaron Singer, and Stuart Shieber. 2020.
\newblock \href
  {https://proceedings.neurips.cc/paper_files/paper/2020/file/92650b2e92217715fe312e6fa7b90d82-Paper.pdf}
  {Investigating gender bias in language models using causal mediation
  analysis}.
\newblock In \emph{Advances in Neural Information Processing Systems},
  volume~33, pages 12388--12401. Curran Associates, Inc.

\bibitem[{Wang et~al.(2023)Wang, Variengien, Conmy, Shlegeris, and
  Steinhardt}]{wang2023interpretability}
Kevin~Ro Wang, Alexandre Variengien, Arthur Conmy, Buck Shlegeris, and Jacob
  Steinhardt. 2023.
\newblock \href {https://openreview.net/forum?id=NpsVSN6o4ul} {Interpretability
  in the wild: A circuit for indirect object identification in {GPT}-2 small}.
\newblock In \emph{The Eleventh International Conference on Learning
  Representations}.

\bibitem[{Warstadt et~al.(2023)Warstadt, Mueller, Choshen, Wilcox, Zhuang,
  Ciro, Mosquera, Paranjabe, Williams, Linzen, and
  Cotterell}]{warstadt-etal-2023-findings}
Alex Warstadt, Aaron Mueller, Leshem Choshen, Ethan Wilcox, Chengxu Zhuang,
  Juan Ciro, Rafael Mosquera, Bhargavi Paranjabe, Adina Williams, Tal Linzen,
  and Ryan Cotterell. 2023.
\newblock \href {https://doi.org/10.18653/v1/2023.conll-babylm.1} {Findings of
  the {B}aby{LM} challenge: Sample-efficient pretraining on developmentally
  plausible corpora}.
\newblock In \emph{Proceedings of the BabyLM Challenge at the 27th Conference
  on Computational Natural Language Learning}, pages 1--34, Singapore.
  Association for Computational Linguistics.

\bibitem[{White et~al.(2021)White, Pimentel, Saphra, and
  Cotterell}]{white-etal-2021-non}
Jennifer~C. White, Tiago Pimentel, Naomi Saphra, and Ryan Cotterell. 2021.
\newblock \href {https://doi.org/10.18653/v1/2021.naacl-main.12} {A non-linear
  structural probe}.
\newblock In \emph{Proceedings of the 2021 Conference of the North American
  Chapter of the Association for Computational Linguistics: Human Language
  Technologies}, pages 132--138, Online. Association for Computational
  Linguistics.

\bibitem[{Wiegreffe et~al.(2024)Wiegreffe, Tafjord, Belinkov, Hajishirzi, and
  Sabharwal}]{wiegreffe-etal-2024-answer}
Sarah Wiegreffe, Oyvind Tafjord, Yonatan Belinkov, Hannaneh Hajishirzi, and
  Ashish Sabharwal. 2024.
\newblock \href {https://arxiv.org/abs/2407.15018} {Answer, assemble, ace:
  Understanding how transformers answer multiple choice questions}.
\newblock \emph{Preprint}, arXiv:2407.15018.

\bibitem[{Wilcox et~al.(2021)Wilcox, Vani, and
  Levy}]{wilcox-etal-2021-targeted}
Ethan Wilcox, Pranali Vani, and Roger Levy. 2021.
\newblock \href {https://doi.org/10.18653/v1/2021.acl-long.76} {A targeted
  assessment of incremental processing in neural language models and humans}.
\newblock In \emph{Proceedings of the 59th Annual Meeting of the Association
  for Computational Linguistics and the 11th International Joint Conference on
  Natural Language Processing (Volume 1: Long Papers)}, pages 939--952, Online.
  Association for Computational Linguistics.

\bibitem[{Wolf et~al.(2020)Wolf, Debut, Sanh, Chaumond, Delangue, Moi, Cistac,
  Rault, Louf, Funtowicz, Davison, Shleifer, von Platen, Ma, Jernite, Plu, Xu,
  Scao, Gugger, Drame, Lhoest, and Rush}]{wolf2020huggingface}
Thomas Wolf, Lysandre Debut, Victor Sanh, Julien Chaumond, Clement Delangue,
  Anthony Moi, Pierric Cistac, Tim Rault, Rémi Louf, Morgan Funtowicz, Joe
  Davison, Sam Shleifer, Patrick von Platen, Clara Ma, Yacine Jernite, Julien
  Plu, Canwen Xu, Teven~Le Scao, Sylvain Gugger, Mariama Drame, Quentin Lhoest,
  and Alexander~M. Rush. 2020.
\newblock \href {https://arxiv.org/abs/1910.03771} {Huggingface's transformers:
  State-of-the-art natural language processing}.
\newblock \emph{Preprint}, arXiv:1910.03771.

\end{thebibliography}

\appendix
\section{Attribution Patching with Integrated Gradients (AtP-IG)}\label{app:atp-ig}
\CR{As described in \S\ref{sec:SAEs}, computing the indirect effect of all features in exact form is computationally expensive, as the number of required forward passes  scales linearly with respect to the number of features in the model. Thus, we employ a linear approximation, $\hat{\text{IE}}$, where the number of required forward and backward passes scales in constant time with respect to the number of features. AtP is defined in Eq.~\ref{eq:atp}. Here, we describe how this is extended to AtP-IG for more accurate estimations.}

\CR{The primary difference between AtP and AtP-IG is that we average the gradient across $K$ intermediate values of $f$ between $a$ and a baseline value $a'$. In our experiments, the baseline is always $0$.\footnote{Note that $a'$ need not be $0$. It could also be taken from a counterfactual input $x'$ where the output behavior of the model differs.} We use $K=10$ in our experiments. The procedure is defined as follows:
\begin{equation}
    \hat{\text{IE}} = (a-a') \cdot \frac{1}{K}\sum_{k=0}^K \frac{\partial m(a' + \frac{k}{K}\cdot (a - a'))}{\partial a}\Big|_x.
    \label{eq:atp-ig}
\end{equation}
That is, given input $x$ and a pre-computed baseline value $a'$, we compute $\frac{\partial d}{\partial a}$ at $K$ intermediate points. At each intermediate point, we intervene on $a$, replacing its activation with what it would have been at that intermediate point. Using this new activation, we recompute $m$, and backpropagate from that to obtain a new gradient value. We take the mean over these gradient values to obtain a more accurate estimate of the slope of $m$ w.r.t.\ $a$. This average slope is then multiplied by the change in $a$ as before.}

\section{Notes on Behavioral Experiments}\label{app:behavioral}
In this paper, we measure the probability assigned by the model to the garden-path and non-garden-path readings via $p($GP$)$ and $p($non-GP$)$, the probability of two individual tokens. Using this sort of naturalistic setup, instead of e.g. prompting the model to explicitly choose one of the garden path sentence's readings is a way to reduce task demands and more accurately judge pre-trained models' performance \citep{hu-levy-2023-prompting,hu2024auxiliary}. However, past work also indicates that setups that pit two alternatives against each other can yield inconsistent results if alternatives are chosen poorly \citep{newman-etal-2021-refining}.

Our reasons for choosing this setup are twofold. First, the most robust setup, which would involve summing the probabilities of all garden-path and non-garden-path continuations, is very both computationally and technically infeasible. Second, while we could instead measure GP and non-GP via sets of tokens, rather than individual tokens, doing so did not change our experimental results in early trial runs. This is due to the fact that our pre-defined GP and non-GP tokens are already the most probable tokens. While there are some tokens that could be used to expand the non-GP token set, e.g. \emph{is, does, should, could}, defining precisely which tokens should be included is challenging: tokens must be third-person verbs that cannot be interpreted as past participles. With all of this in mind, we stick with a simpler setup.

\section{Faithfulness}\label{app:faithfulness}
Faithfulness is a metric commonly employed in circuit analysis studies \citep[e.g.,][]{wang2023interpretability,conmy2023towards,hanna2023gpt2,miller2024transformer,marks2024sparse,hanna2024have}. The metric aims to capture the proportion of model behavior on dataset $\mathcal{D}$ explained by the circuit. More concretely, given target metric $m$, full model $\mathcal{M}$, and circuit $\mathcal{C}$, we follow \citet{marks2024sparse} in defining faithfulness $F$ as the average normalized ratio of $m$ given $\mathcal{C}$ over $m$ given the full model:
\begin{equation}
    F = \mathbb{E}_{x\in\mathcal{D}}\left[\frac{m(\mathcal{C}, x) - m(\varnothing, x)}{m(\mathcal{M}, x) - m(\varnothing, x)}\right]
\end{equation}
We define $m$ as the logit difference between the garden-path completion and the non-garden-path completion given $x$. $m(\varnothing)$ refers to the logit difference when ablating \emph{all} features. Here, an ablation entails setting a feature's activation to $0$ before reconstructing the activations. The intuition is that the circuit should capture the same proportion of $m$ above its prior (i.e., in the absence of any input-specific information) than the full model captures for as many examples as possible.

Note that when computing faithfulness, we include all nodes whose \emph{absolute} $\hat{\text{IE}}$ values surpass the threshold. This means that we include positive-$\hat{\text{IE}}$ components that increase the difference in favor of non-garden-path continuations, and negative$\hat{\text{IE}}$ components that increase the difference in favor of garden-path continuations. This is because, in ambiguous settings, both readings are possible, and we would like to recover features that are sensitive to both readings.

When computing faithfulness, \citet{marks2024sparse} give approximately the first $\frac{1}{4}$ of the layers in the model for free---that is, all features in the embedding layer and through the end of layer 1 for Pythia. In other words, all features in these layers are implicitly included in the circuit, regardless of whether they passed the effect threshold. The reasoning is that these features are generally only responsible for detecting that certain tokens have appeared in the inputs; thus, without them, the model would not be aware that these tokens have appeared, and it would therefore not be possible to perform the task. Unlike in their setting, we do not have a distinction between the circuit discovery setting and the evaluation setting,\footnote{As there is no optimization involved in obtaining the circuit, a held-out set is not always used in circuit discovery. That said, we acknowledge that evaluating circuits on held-out data makes it more likely that the discovered mechanism will generalize to wider distributions of inputs.} but we do find that many embedding and layer-0 features still do not appear in the circuits that should. These generally correspond to word detectors for tokens that only appeared in one example in $\mathcal{D}$. Thus, for Pythia, we give the model only the embedding and layer-0 features for free when computing faithfulness. For Gemma 2, we find that layers 0--2 contain word detector features, so we give all features layers 0, 1, and 2 for free when computing faithfulness.

Our faithfulness results for Pythia-70m in \S\ref{subsec:feature} are either much lower or much higher than 1.0. For NP/S, we obtain a faithfulness of $0.20$, which means that we have recovered 20\% of the logit difference between the non-garden-path and garden-path continutions as compared to the full model. For NP/Z, we obtain a faithfulness of $3.48$, meaning that our circuit's logit difference is over 300\% higher than the full model's. For Gemma-2-2b (circuits in App.~\ref{app:full-feature-circuit}), the NP/S circuit has faithfulness $0.07$, whereas the NP/Z circuit has faithfulness $0.23$. $0.20$ is on par with the faithfulness values of \citet{marks2024sparse} for subject--verb agreement, but $3.48$ is very high, and likely means that we have not captured many of the important \emph{negative-effect} (garden-path-upweighting) features. Indeed, when we lower the effect threshold, we observe that faithfulness slowly (but non-monotonically) approaches 1. The Gemma NP/S circuit's low faithfulness of $0.07$ suggests that we must include many more features to capture the full mechanism. This is unsuprising, given that this model is significantly larger than Pythia and should therefore require more features to achieve the same behavior.

In follow-up analyses, we find that achieving close to a faithfulness of 1 requires many hundreds of features for Pythia-70m---and thousands for Gemma-2-2b.\footnote{See App.~\ref{app:other-models} for faithfulness values for Gemma 2's circuits.} Currently, this number of features is not tractable to annotate manually, and our initial experiments revealed that automated feature labeling methods such as those of \citet{bills2023language} tend to not be sensitive to syntactic distributions, instead preferring purely lexical or semantic interpretations of feature activation patterns. Future work could enable new mechanistic analyses by improving the ability of automated neuron/feature explanation techniques to detect syntactic distributional features.

\section{Causal Experiments on a Larger Dataset}\label{app:large-dataset}

\begin{figure}
    \centering
    \includegraphics[width=\linewidth]{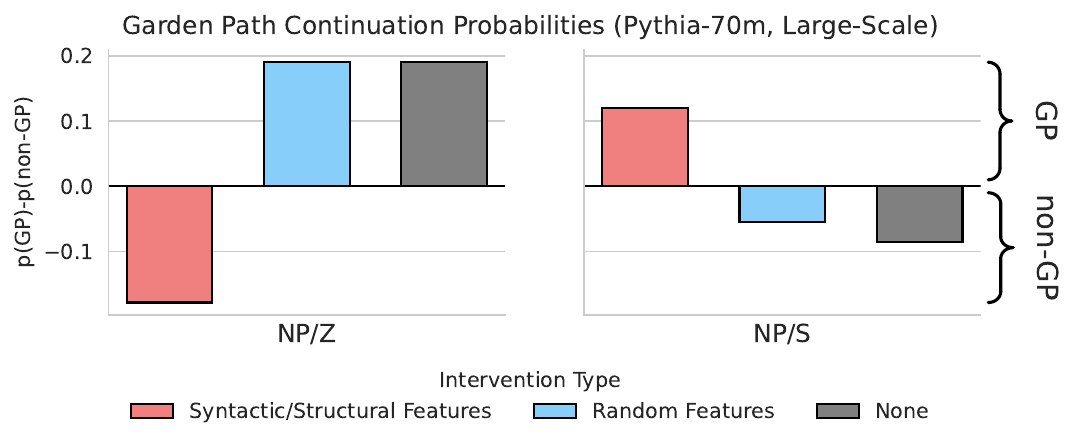}
    \caption{Mean difference in probability of GP and non-GP continuations under interventions for Pythia-70m, on the larger-scale SAP Benchmark. Error bars indicate the standard error of the mean (SEM); however, SEM is near-0 and thus not visible. Interventions on interpretable features reverse model behavior, as expected; random interventions do not change behavior.}
    \label{fig:causal_pythia-70m_largescale}
\end{figure}

The dataset from \citet{arehalli-etal-2022-syntactic} that we adapt for feature circuit finding is small. This is important, because we manually adapt it to be compatible with our methods. In particular, we craft the unambiguous examples described in \S\ref{subsec:behavioral}, and also force every example to have the same token length. The latter is key, because, if we wish to estimate the importance of a feature \emph{at a given position} over a number of different examples, each example must have the same token length, and the type of token at each position (e.g. verb, final noun, etc.) must be the same. For this reason, it is currently infeasible to run these experiments (or any other feature experiments) on a larger, non-handcrafted dataset.

These same restrictions do not apply to the causal experiment. In that experiment, as long as we know where the verb and final noun are located in the sentence, our sentences may have different lengths, and different semantic content at each position. Taking advantage of this, we run our causal experiment (see \S\ref{subsec:causal}) again on a larger dataset. We use the syntactic ambiguity benchmark (SAP Benchmark, \citealp{huang2024large}), of which \citeauthor{arehalli-etal-2022-syntactic}'s dataset is a subset. This dataset has 7952 NP/Z sentences and 7948 NP/S sentences. We follow the methods from \S\ref{subsec:causal} exactly, taking special care to accommodate the different lengths and positions in this dataset. We perform this analysis only on Pythia-70m-deduped; performing this on Gemma-2-2b would be rather slow.

Our results (\Cref{fig:causal_pythia-70m_largescale}) show that the features we found in \S\ref{subsec:feature} generalize to this larger dataset as well, even though they were found on a very small subset thereof. Our ablations successfully induce the model to produce non-GP continuations for NP/Z sentences, and GP continuations for NP/S sentences, reversing its initial preferences, exactly as in \S\ref{subsec:causal}. Again, the random ablations are ineffective, leaving performance close to the no-intervention baseline.

\section{Results for Gemma-2-2b}\label{app:other-models}
To ensure our findings are not merely a function of model size or the Pythia SAEs, we also replicate the experiments for Gemma-2-2b. We first present results for the behavioral analysis (App.~\ref{subsec:gemma-behavioral}). Then, after discovering feature circuits for NP/S and NP/Z (shown in App.~\ref{app:full-feature-circuit}), we causally verify the labels we assign to these features (App.~\ref{subsec:gemma-causal}). \CR{We use \citeposs{lieberum-etal-2024-gemma} Gemma-2-2b SAEs with 16,384 features.}

\subsection{Behavioral experiments}\label{subsec:gemma-behavioral}

\begin{figure}[t]
    \includegraphics[width=\linewidth]{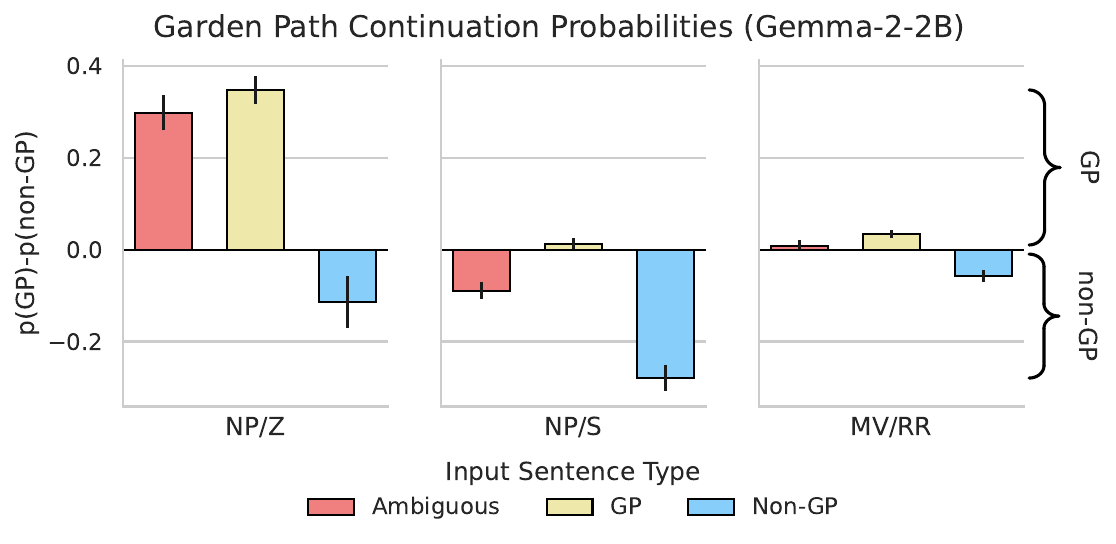}
    \caption{Mean difference in probability of tokens corresponding to garden path (``,''/``.'') and non-garden-path (``was'') readings of the input for Gemma-2-2b, aggregated by garden path structure. Error bars indicate the standard error of the mean. Inputs are either \amb{ambiguous}, or compatible with only a \gp{garden-path} or \nongp{non-garden-path} reading. (Non)-garden-path tokens are more probable given (non)-garden-path inputs. In ambiguous cases, the model prefers the garden path reading, except for NP/S inputs.}
    \label{fig:behavioral_gemma}
\end{figure}
Here, we present behavioral results for Gemma-2-2b (Figure~\ref{fig:behavioral_gemma}). The experimental setup is the same as that described in \S\ref{subsec:behavioral}. For all sentence structures, findings are largely consistent as those for Pythia: Gemma 2 upweights and downweights garden-path tokens in appropriate contexts. For ambiguous inputs, the model gives more probability to garden-path continuations in NP/Z, but non-garden-path continuations in NP/S. For MV/RR, Gemma-2-2b assigns higher probability to non-GP continuations than GP continuations in contexts that license non-GP continuations only. This is distinct from what was observed in Pythia, where probabilities for both continuations were closer to each other, with GP continuations being slightly more probable.

\begin{figure}
    \includegraphics[width=\linewidth]{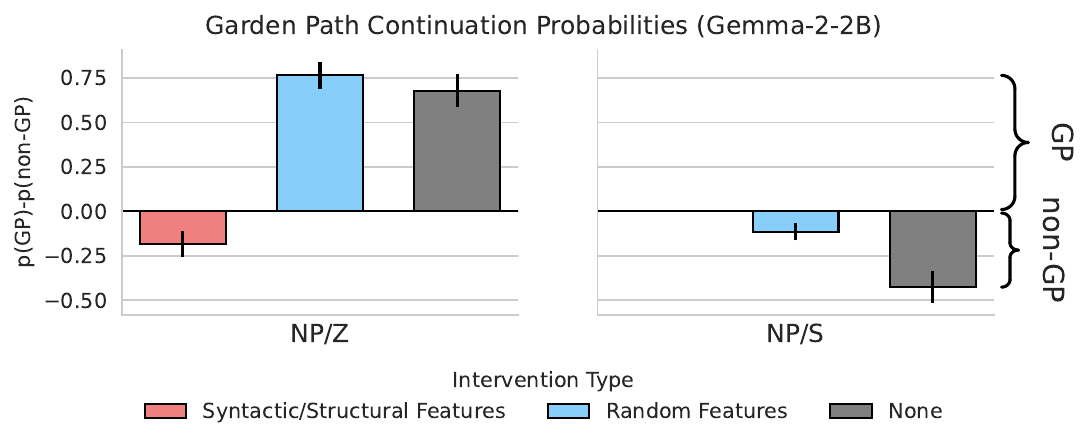}
    \caption{Difference in mean probability of tokens corresponding to garden-path and non-garden-path readings of the input for Gemma-2-2b, aggregated by garden path structure. Error bars indicate the standard error of the mean. We either intervene on interpretable features to induce the opposite behavior, or intervene on random features. The interventions on interpretable features are effective in the way we expect, whereas random interventions do not change behavior.}
    \label{fig:gemma-causal}
\end{figure}

\begin{figure*}
    \includegraphics[width=\linewidth]{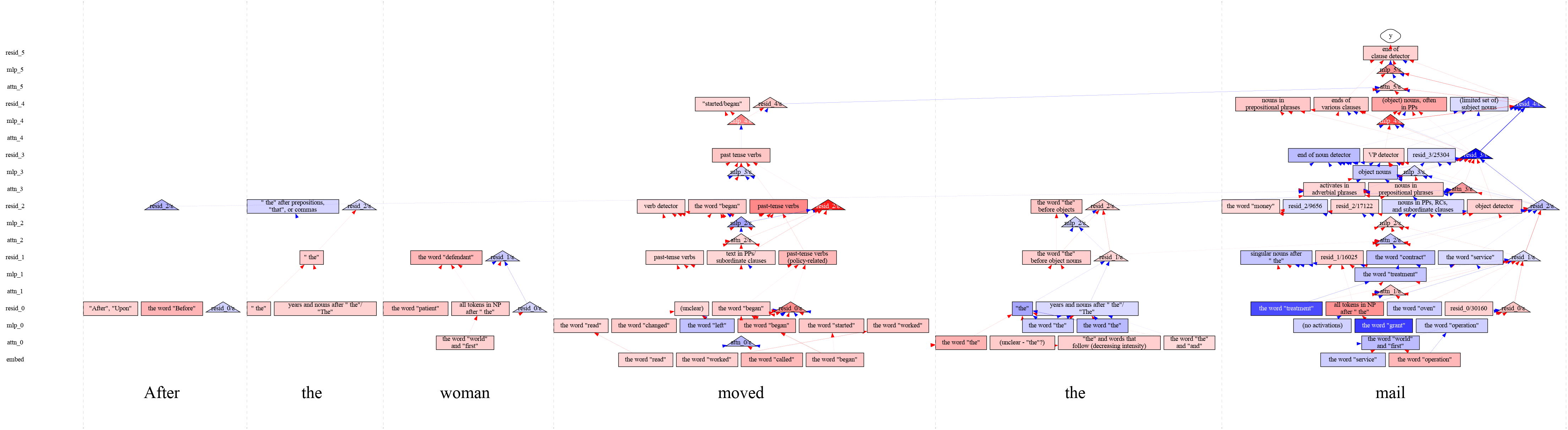}
    \caption{Sparse feature circuit for Pythia-70m on the NP/Z garden path structure. Features with larger positive effects are colored in deeper shades of blue; features with larger negative effects are colored in deeper shades of red. Zoom in to view feature annotations.}
    \label{fig:sfc_npz_pythia}
\end{figure*}

\begin{figure*}
    \includegraphics[width=\linewidth]{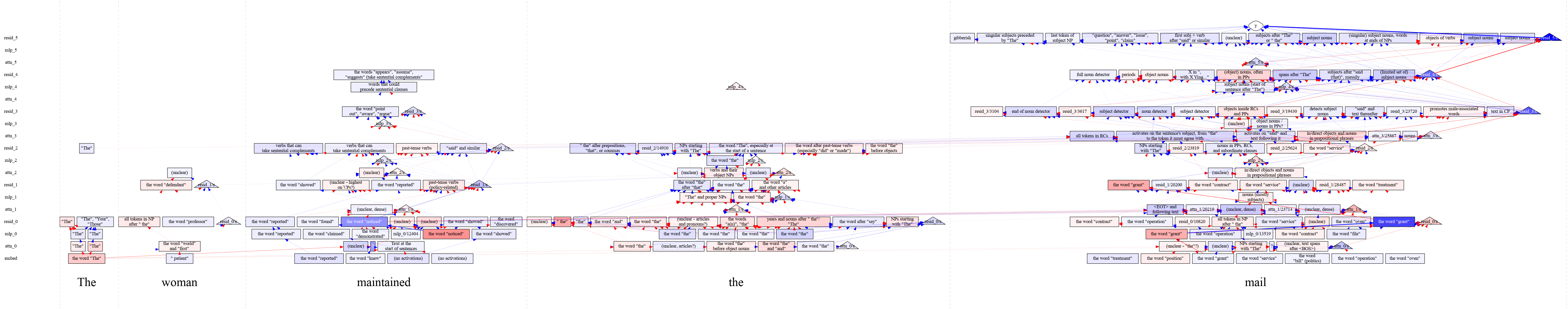}
    \caption{Sparse feature circuit for Pythia-70m on the NP/S garden path structure. Features with larger positive effects are colored in deeper shades of blue; features with larger negative effects are colored in deeper shades of red. Zoom in to view feature annotations.}
    \label{fig:sfc_nps_pythia}
\end{figure*}

\subsection{Causal verification}\label{subsec:gemma-causal}
Having shown that Gemma 2 prefers the GP reading for NP/Z, we aim to induce the dispreferred non-GP reading by clamping subject detectors to high activations (100.0) at the final noun, and clamping object detectors to low activations (0.0). Note that the artificial high activation here is much larger (100.0) for Gemma 2 than what we used for Pythia (2.0). This is because the activations are generally much larger in the Gemma 2 SAEs; indeed, activations of 100.0 are not necessarily out of distribution. For NP/S, Gemma 2 prefers to non-GP reading, so we attempt to induce the dispreferred GP reading by doing the opposite---namely, setting the subject detector and object detector features to 0.0 or 100.0, respectively, and by clamping the sentential-clause-verb detectors to 0.0 at the verb position. As in \S\ref{subsec:causal}, we compare to a baseline where we clamp the same number of randomly sampled features to high or low activations.

Our findings (\Cref{fig:gemma-causal}) suggest that the features we find are causally relevant, and in the way we expect. For NP/Z, we can change the model's probabilities such that $p(\text{non-GP}) > p(\text{GP})$. For NP/S, we can decrease the originally preferred $p(\text{non-GP})$. The increases in $p(\text{GP})$ is difficult to visualize, but present: $p(\text{GP})$ is increased from $5\times 10^{-6}$ to $4\times 10^{-3}$, and because the new $p(\text{non-GP})$ is $1\times 10^{-9}$, we have induced a relative preference for the originally dispreferred reading. Nonetheless, it is likely that other continuations outside of the GP and non-GP tokens we consider have now become more probable than either of these two possibilities.

\section{Feature Circuits}\label{app:full-feature-circuit}
Here, we present the full sparse feature circuits for NP/S and NP/Z. We include feature circuits for both Pythia-70m and Gemma-2-2b. For both Pythia circuits, we set the node threshold to $0.1$ and the edge threshold to $0.001$. To keep the feature circuit a size that will fit onto a page (and to keep the number of features we must manually annotate reasonable), we slightly increase the node threshold to $0.12$ when discovering the Gemma 2 circuits.

Because we include any node where the \emph{absolute value} of the $\hat{\text{IE}}$ is over the node threshold, we include positive- and negative-effect features. Positive-effect features increase the relative probability of the non-garden-path continuation over the garden-path-continuation, whereas negative-effect features increase the garden-path continuation probability relative to the non-garden-path continuation. We manually annotate all features in these circuits by observing their activation patterns and the tokens whose probabilities are most affected when the feature is ablated.\footnote{We acknowledge that there are issues in both precision and recall when assigning textual explanations to neurons \citep{huang-etal-2023-rigorously}, and that these issues extend to sparse features. Our causal verification experiments mitigate this somewhat, but natural language is ultimately an ambiguous medium for expressing the functional  role of model components. Future work should consider more formal ways of describing sparse features.}

\begin{figure*}
    \includegraphics[width=\linewidth]{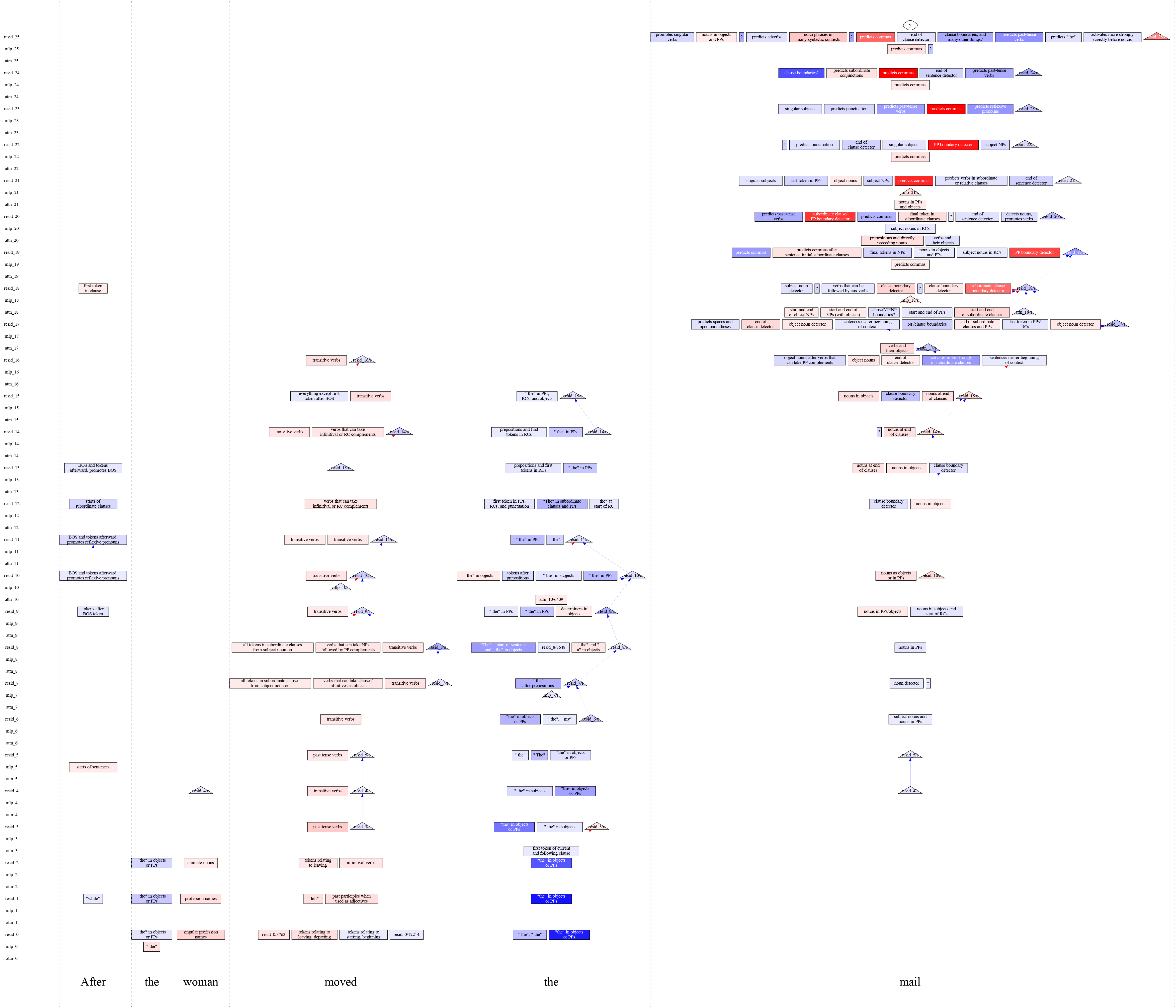}
    \caption{Sparse feature circuit for Gemma-2-2b on the NP/Z garden path structure. Features with larger positive effects are colored in deeper shades of blue; features with larger negative effects are colored in deeper shades of red. Zoom in to view feature annotations.}
    \label{fig:sfc_npz_gemma2}
\end{figure*}

\begin{figure*}
    \includegraphics[width=\linewidth]{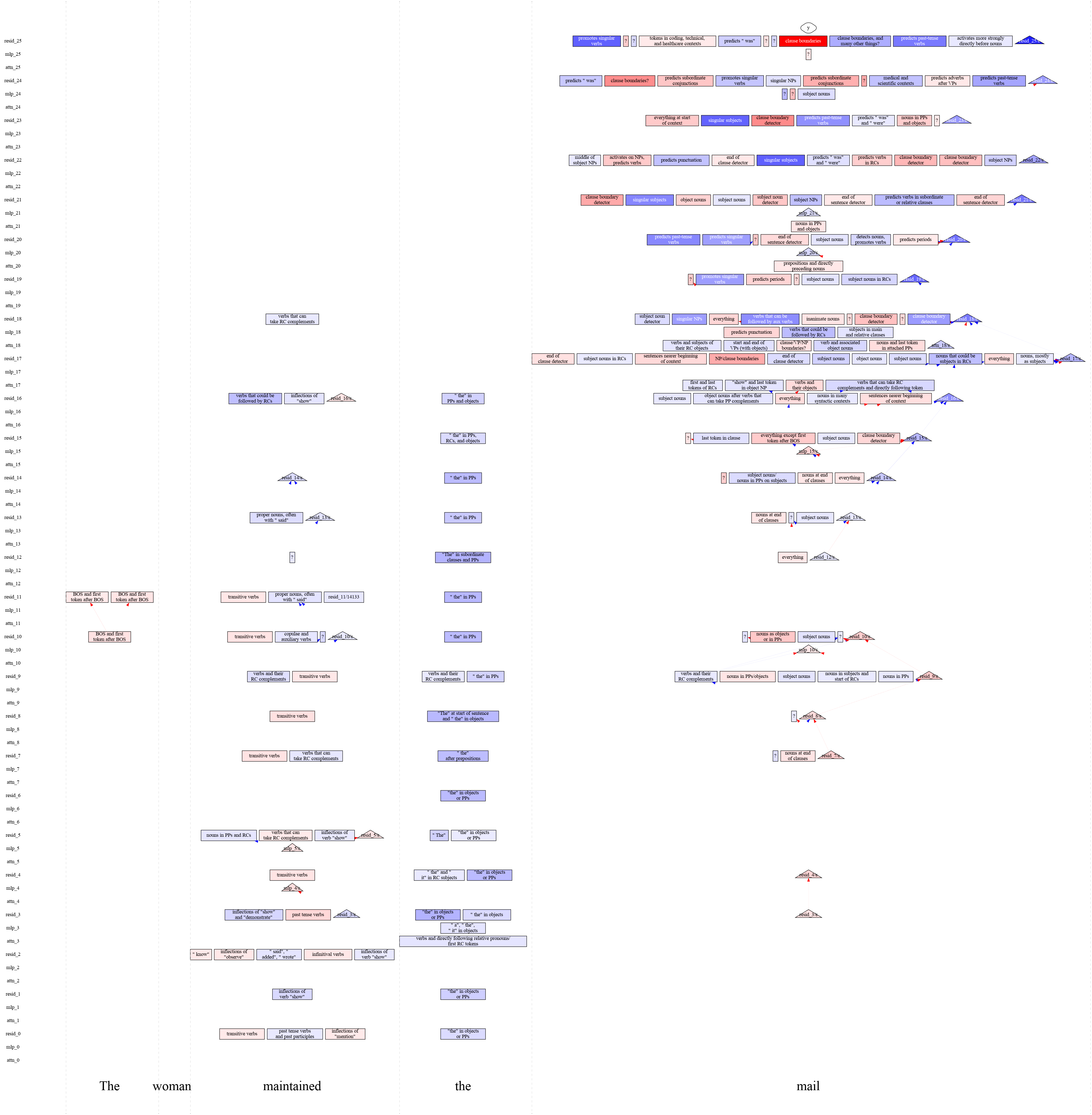}
    \caption{Sparse feature circuit for Gemma-2-2b on the NP/S garden path structure. Features with larger positive effects are colored in deeper shades of blue; features with larger negative effects are colored in deeper shades of red. Zoom in to view feature annotations.}
    \label{fig:sfc_nps_gemma2}
\end{figure*}

The sparse feature circuits for NP/Z (\Cref{fig:sfc_npz_pythia,fig:sfc_npz_gemma2}) are similar across models. Both contain primarily spurious or word-level features in the lower layers, and more syntax-sensitive features in the upper layers. See Figure~\ref{fig:circuit-diagram} for a condensed version of Pythia's NP/Z circuit, where we summarize the main categories of features and their effects on the model's preferred continuation. Pythia's NP/Z circuit contains 65 features, and Gemma 2's contains 182.

\begin{figure}
\centering
\includegraphics[width=\linewidth]{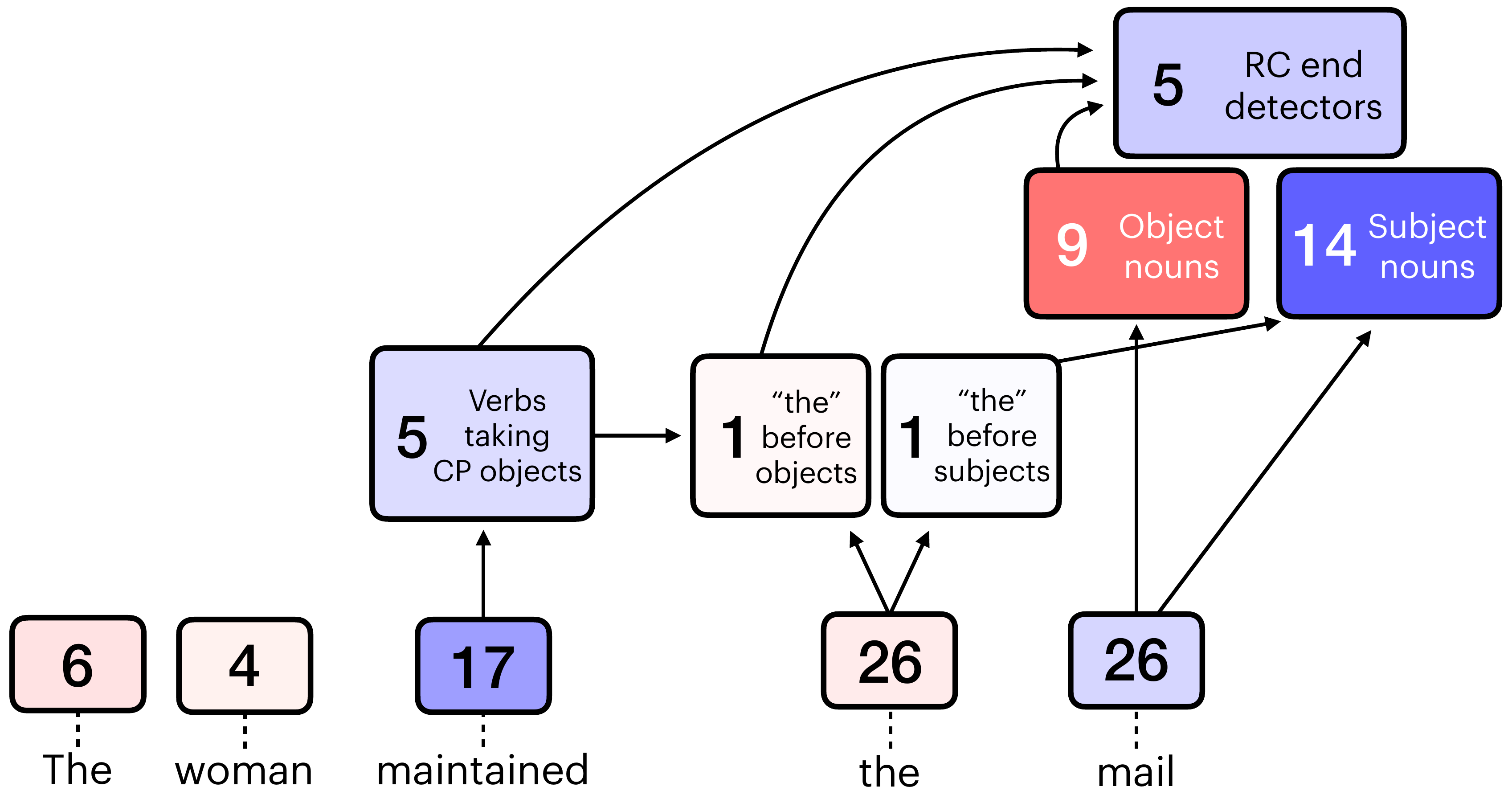}
\caption{Pythia-70m's feature circuit for processing NP/S garden path sentences. We group features by their functional role in the circuit and display the number of features in each group. Red features have negative scores, and push the model towards the garden path reading; blue features have positive scores, and do the opposite. Unlabeled early-layer features are word detectors. Many late-layer features encode syntactic features, whereas early-layer features largely consist of word detectors and heuristics. Note that we exclude features that are difficult to interpret from the feature counts.} \label{fig:circuit-diagram-nps}
\end{figure}

The sparse feature circuits for NP/S (\Cref{fig:sfc_nps_pythia,fig:sfc_nps_gemma2}) show similar trends. See Figure~\ref{fig:circuit-diagram-nps} for a condensed version of Pythia's NP/S circuit. Note that more of the features have negative effects in the NP/Z circuits than in the NP/S circuits, as both models more strongly prefer the garden path continuations for NP/Z inputs. Pythia's NP/S circuit contains 155 features, and Gemma 2's contains 179.

\section{Structural Probe Training and Results}\label{app:structural-probe}

\subsection{Probe Details}
We use \citeposs{eisape-etal-2022-probing} MLP action probes to probe Pythia-70m's internal parse information. These probes take in the representations of two words,\footnote{Our dataset contains no multi-token words, but during training, multi-token words are aggregated to form a single-token representation.} and compute the probability of a given parse action $a$ as
\begin{equation}
    P(a)\propto\exp\left(e^\top_a\text{MLP}([\mathbf{h}_1, \mathbf{h_2}])+b_a\right),
\end{equation}

where $\mathbf{h}_1$ and $\mathbf{h_2}$ are the hidden representations of the words whose relation you wish to predict, and $e_a$ and $b_a$ are learned weight and bias terms respectively.

While we consider the parse probes in isolation, \citet{eisape-etal-2022-probing} use them as part of a larger parsing architecture. Specifically, they rely on the arc-standard dependency formalism \citep{nivre2004incrementality}, which parses the input into subtrees which are placed on a stack and repeatedly combined with each other via parse actions in order to obtain a full (incremental) parse of the input. 

There are three parse actions: \texttt{LEFT-ARC}, which pops the first two subtrees $\mathbf{s}_1,\mathbf{s}_2$ off the stack and draws an arc from $\mathbf{s}_1$ to $\mathbf{s}_2$; \texttt{RIGHT-ARC}, which does the same, but draws the opposite arc; and \texttt{GEN}, which indicates no relation, and moves the parsing process forward by generating another token. 

Notably, \texttt{LEFT-ARC} and \texttt{RIGHT-ARC} refer not to the position of words in the sentence, but to the direction of the arc between popped subtrees. This is why the non-garden-path reading corresponds to the \texttt{LEFT-ARC} action; during normal parsing of our garden path fragments, the final noun heads $\mathbf{s}_1$, while the verb heads $\mathbf{s}_2$, so arcs are reversed with respect to their appearance on paper.

\begin{figure}
    \centering
    \includegraphics[width=\linewidth]{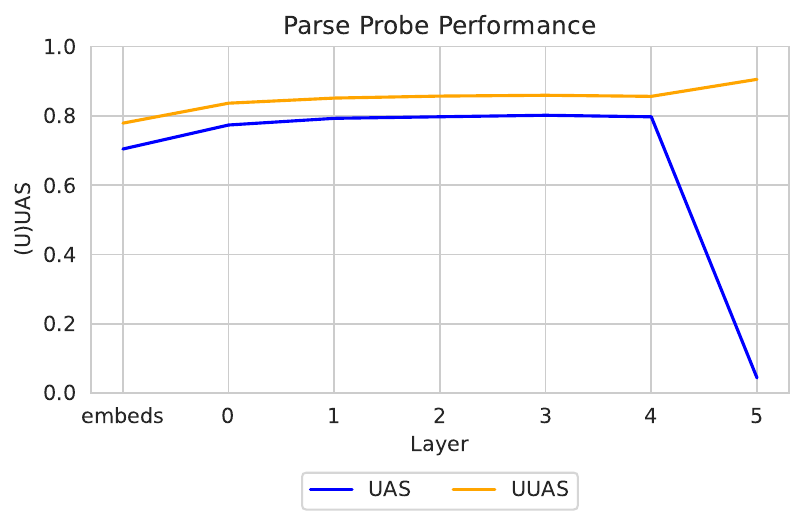}
    \caption{Probe unlabeled attachment score (UAS) and undirected unlabeled attachment score (UUAS) on the Penn Treebank test split, by layer.}
    \label{fig:probe-uas}
\end{figure}

\subsection{Probe Training}
Following \citet{eisape-etal-2022-probing}, we train our probes on the training split of the Penn Treebank \citep{taylor2003penn};\footnote{Note that the Penn Treebank does not originally come with these splits, which were defined in \citet{hewitt-manning-2019-structural}. Documents 2-21 of the WSJ portion of the dataset are considered the train split; document 22 is the validation split; document 23 is the test split.} we use essentially the same hyperparameters as in their work, modified to work with Pythia-70m-deduped, rather than GPT-2. Then, we also record unlabeled attachment score (UAS) and undirected unlabeled attachment score (UUAS) on the test split, in order to verify that our probes are effective.

Our results (\Cref{fig:probe-uas}) show that the probes are indeed effective. The probes' UAS and UUAS are similar to the values. The UAS for the last layer is unusually low, even considering the last layer's lower performance in \citet{eisape-etal-2022-probing}, indicating that the direction of dependency relations is not captured, but this tracks with the probes' poor performance on garden path sentences using representations from that layer.

\subsection{Probe Evaluation on Unambiguous Garden-Path-Derived Stimuli}\label{app:probe-unambiguous}
\begin{figure}
    \centering
    \includegraphics[width=\linewidth]{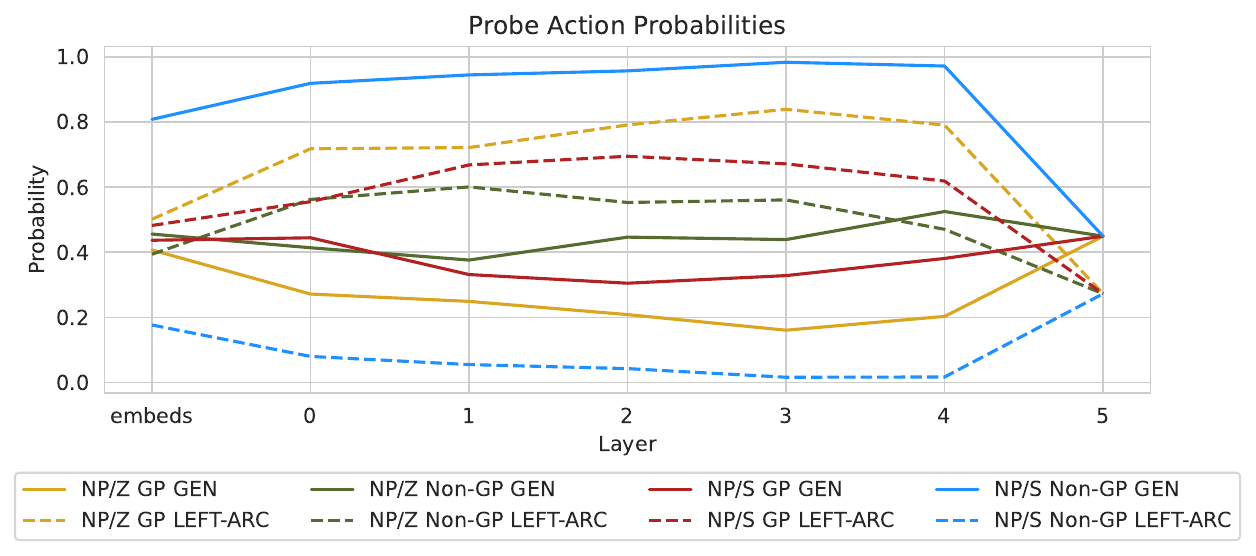}
    \caption{Probe action probability across layers and sentence types (GP and non-GP). \texttt{GEN} corresponds to the non-GP reading, and \texttt{LEFT-ARC} to the GP reading (\texttt{RIGHT-ARC} is implausible and is excluded, as it always receives low probability). GP sentences elicit primarily \texttt{LEFT-ARC}; non-GP sentences elicit \texttt{GEN}. However, both readings do have non-zero probability in both cases.}
    \label{fig:probe-behavior-comparison}
\end{figure}

In \S\ref{subsec:probes}, we found that the probes' judgments on regarding the parse of the sentence matched with our observations based on features and behavior. But do these probes also behave sensibly on stimuli whose parse is known? To test this, we evaluate the probes on the unambiguous stimuli from our dataset (\Cref{tab:examples}), and record their action probabilities. Ideally, the probes should prefer \texttt{LEFT-ARC} on GP sentences, and \texttt{GEN} on non-GP sentences.

Our results (\Cref{fig:probe-behavior-comparison}) show that this is generally the case: GP sentences elicit primarily \texttt{LEFT-ARC}; non-GP sentences elicit \texttt{GEN}. However, the model struggles on NP/Z non-GP sentences, perhaps because these are the least plausible ones; such sentences are generally written with a comma after the verb, and read strangely. Moreover, both readings do have non-zero probability in most cases, even though their construction should preclude the alternative reading. The probes thus seem somewhat less attuned to syntactically in/valid readings than LM probabilities are.

\subsection{Feature Consistency}\label{app:probe-feature-consistency}
\begin{figure}
    \centering
    \includegraphics[width=\linewidth]{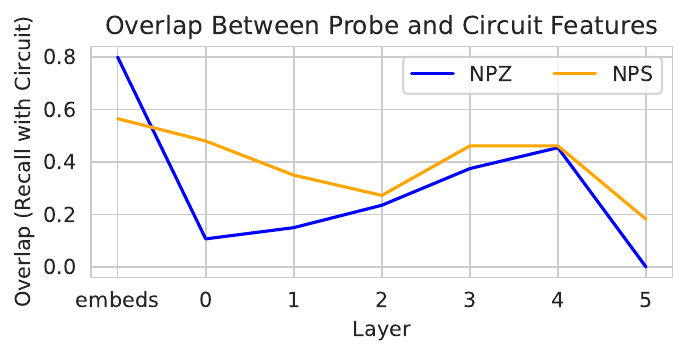}
    \caption{Overlap of probe and circuit features, computed as recall with respect to circuit features. Random chance overlap is near 0, but probe features overlap significantly with circuit features.}
    \label{fig:probe-features}
\end{figure}

We can test the consistency between whole-model and probing methods by performing feature analysis with our structural probe. Each probe takes as input residual stream activations, for which we have SAEs; we can thus use AtP-IG to find features that influence the quantity $p($\texttt{LEFT-ARC}$)$ - $p(\texttt{GEN})$, just as we previously found model features that influenced $p($GP$)$ - $p($non-GP$)$. For each structure (NP/Z and NP/S) and layer of the model, we take $F_c$, the set of features in that layer of the circuit, and $F_p$, the set containing the top-$|F_c|$ features for the probe. We quantify the sets' overlap via recall, $\frac{|F_c\cap F_p|}{|F_c|}$. The expected recall for random features would be very near 0; however, \Cref{fig:probe-features} shows that the probe features' recall is quite high. This overlap is highest (0.6-0.8) in the embeddings, but there is also high overlap (0.35-0.45) in layers 3 and 4, which contain interpretable, high-level syntactic features. Thus, even though these probes were trained and attribution performed in very ways, the same underlying features are responsible.

\begin{table}
    \centering
    \resizebox{\linewidth}{!}{
    \begin{tabular}{lrrr}
        \toprule
        Model & MV/RR & NP/S & NP/Z \\
        \midrule
        Gemma-2-2b & 73.0 & 83.3 & 70.9 \\
        \ \ --Neg MV/RR features & 73.0 & 83.3 & 79.2 \\
        \ \ --Neg NP/S features & 70.9 & 83.3 & 70.9 \\
        \ \ --Neg NP/Z features & 73.0 & 85.4 & 72.9 \\
        \midrule
        \ \ --Pos MV/RR features & 70.9 & 83.3 & 60.4 \\
        \ \ --Pos NP/S features & 70.9 & 83.3 & 75.0 \\
        \ \ --Pos NP/Z features & 73.0 & 83.3 & 70.9 \\
        \bottomrule
    \end{tabular}}
    \caption{Accuracies on follow-up reading comprehension questions given garden path sentences. ``Pos'' refers to ablating positive-effect features, or those promoting the non-garden-path reading. ``Neg'' refers to ablating negative-effect features, or those promoting the garden-path reading. Performance generally changes little under ablations, except for NP/Z when ablating MV/RR features.}
    \label{tab:qa_results_appendix}
\end{table}

\section{Reading Comprehension Questions: Performance under ablations}
Here, we assess the extent to which we can influence model performance in garden path reading comprehension questions by ablating the GP-promoting or non-GP-promoting features. Using the same dataset as in \S\ref{sec:reanalysis-repair}, we ablate the top 10 and bottom 10 features discovered from \S\ref{subsec:feature} and then remeasure performance. We hypothesize that ablating the positive features (those promoting the non-garden-path reading) will cause performance to drop, whereas ablating the negative features (those promoting the garden-path reading) will cause performance to increase.

Our results (Table~\ref{tab:qa_results_appendix}) indicate that the ablations are largely ineffective at changing behavior. In some cases, performance does decrease or increase, but typically not to a significant extent. Where differences are significant, it is generally not for the structure from which the features were discovered. For example, ablating positive MV/RR features causes a significant increase in performance for NP/Z questions, and ablating negative MV/RR features also increases performance on NP/Z questions.

\section{Data Artifacts, Experimental Details, and Risks}
\paragraph{Data Artifacts} In this paper, we mainly use \citeposs{arehalli-etal-2022-syntactic} garden path sentence dataset, which is in turn a subset of the syntactic ambiguity benchmark (SAP, now published as \citealp{huang2024large}), a larger garden path sentence dataset. The latter uses an MIT license, and our use case (intepretability and psycholinguistic research) is appropriate for the license. The other datasets---BoolQ \citep {clark-etal-2019-boolq} and MCQA \citep{wiegreffe-etal-2024-answer}---are released with licenses (CC BY-SA 3.0 and Apache 2.0) compatible with research use. All datasets are entirely in English.

We also craft two follow-up sentences per NP/Z and NP/S sentence in the aforementioned dataset. These follow-up sentences, and code for our experiments, will be released upon acceptance. 

\paragraph{Experimental Details} We perform our experiments using an Nvidia A100 (80GB) GPU and Nvidia RTX 6000 Ada GPU. The former is helpful for finding Gemma feature circuits with a low threshold. In total, running all experiments should take no more than 5 GPU-days on the former (perhaps less). Most of the runtime comes from running the Gemma experiments and training parse probes.

All experiments are implemented in PyTorch \citep{paszke2019pytorch} using the NNsight interpretability framework \citep{fiottokaufman2024nnsight}. All LMs used were accessed via HuggingFace \citep{wolf2020huggingface}.

\paragraph{Risks} Because our study only attempts to interpret pre-trained models, we believe that it poses few risks; similarly, the basic follow-up questions carry with them few risks.

\end{document}